\documentclass[sigconf]{acmart}
\AtBeginDocument{%
  }

\setcopyright{acmlicensed}
\copyrightyear{2018}
\acmYear{2018}
\acmDOI{XXXXXXX.XXXXXXX}
\acmConference[Conference acronym 'XX]{Make sure to enter the correct
  conference title from your rights confirmation email}{June 03--05,
  2018}{Woodstock, NY}
\acmISBN{978-1-4503-XXXX-X/2018/06}

\usepackage{hyperref}       
\usepackage{url}            
\usepackage{booktabs}       
\usepackage{amsfonts}       
\usepackage{nicefrac}       
\usepackage{microtype}      
\usepackage{xcolor}         

\usepackage{graphicx}
\usepackage{subcaption}
\usepackage{cuted}

\usepackage{amsmath}
\usepackage{mathtools}
\usepackage{amsthm}
\usepackage{algorithm}
\usepackage{algorithmic}
\usepackage{colortbl} 
\usepackage{pifont}

\usepackage{enumitem}
\usepackage{svg}
\usepackage{booktabs}
\usepackage{multirow}
\usepackage{appendix}
\usepackage{wrapfig}
\usepackage{makecell}
\newcolumntype{C}[1]{>{\centering\arraybackslash}m{#1}}

\usepackage[most]{tcolorbox}
\usepackage{dashrule}
\usepackage{balance}

\usepackage[capitalize,noabbrev]{cleveref}
\crefname{section}{Sec.}{Secs.}
\Crefname{section}{Section}{Sections}
\Crefname{table}{Table}{Tables}
\crefname{table}{Tab.}{Tabs.}

\newcommand{\ours}[0]{Med-R$^3$\ }

\usepackage{caption}
\usepackage{multicol}
\begin{document}

\title{Med-R$^3$: Enhancing Medical Retrieval-Augmented Reasoning of LLMs via Progressive Reinforcement Learning}

\author{Keer Lu$^{\dagger}$, Zheng Liang$^{\mathsection}$, Youquan Li$^{\dagger}$, Jiejun Tan$^{\mathsection}$, Da Pan$^{\mathsection}$, Shusen Zhang$^{\mathsection}$, 
Guosheng Dong$^{\mathsection}$, \\
Bin Cui$^{\dagger}$, Yunhuai Liu$^{\dagger}$, Wentao Zhang$^{\dagger}$ \\
}
\affiliation{
\country{$^\dagger$Peking University, 
$^\mathsection$Baichuan Inc.}
}
\email{{keer.lu, youquan.li}@stu.pku.edu.cn, {bin.cui, yunhuai.liu, wentao.zhang}@pku.edu.cn}

\renewcommand{\shortauthors}{Keer Lu et al.}

\begin{abstract}
In medical scenarios, effectively retrieving external knowledge and leveraging it for rigorous reasoning is important. 
Despite their potential, 
existing work has predominantly focused on enhancing either retrieval or reasoning capabilities in isolation, with little attention given to their joint optimization, 
which leads to limited coordination between the two processes. 
Additionally, current methods rely heavily on supervised fine-tuning (SFT), 
which can cause models to memorize existing problem-solving pathways, thereby restricting their generalization ability when confronted with novel problem contexts. 
Furthermore, while some studies have explored to improve retrieval-augmented reasoning in general domains via reinforcement learning, their reward function designs do not adequately capture the specific demands of the medical domain. 
To address these challenges,  
we introduce \textbf{Med-R$^3$}, a \textbf{Med}ical \textbf{R}etrieval-augmented \textbf{R}easoning framework driven by progressive \textbf{R}einforcement learning. 
In this framework, 
we first develop the model’s ability to perform logical reasoning over medical problems. 
Subsequently, based on this foundation, we adaptively optimize the retrieval capability to better align with the characteristics of knowledge corpus and external information utilization throughout the reasoning process. 
Finally, we conduct joint optimization of the model's retrieval and reasoning coordination. 
Experiments indicate that 
\textbf{Med-R$^3$} 
could achieve state-of-the-art performances, 
with Qwen3-8B + Med-R$^3$ surpassing closed-sourced GPT-4o-mini by 12.22\% at a comparable parameter scale, while Qwen2.5-14B augmented with \ours shows a more substantial gain of 16.31\%. 
\end{abstract}




\received{20 February 2007}
\received[revised]{12 March 2009}
\received[accepted]{5 June 2009}

\addtocontents{toc}{\protect\setcounter{tocdepth}{0}}

\maketitle

\begin{figure}[ht]
    \centering
    \includegraphics[width=1\linewidth]{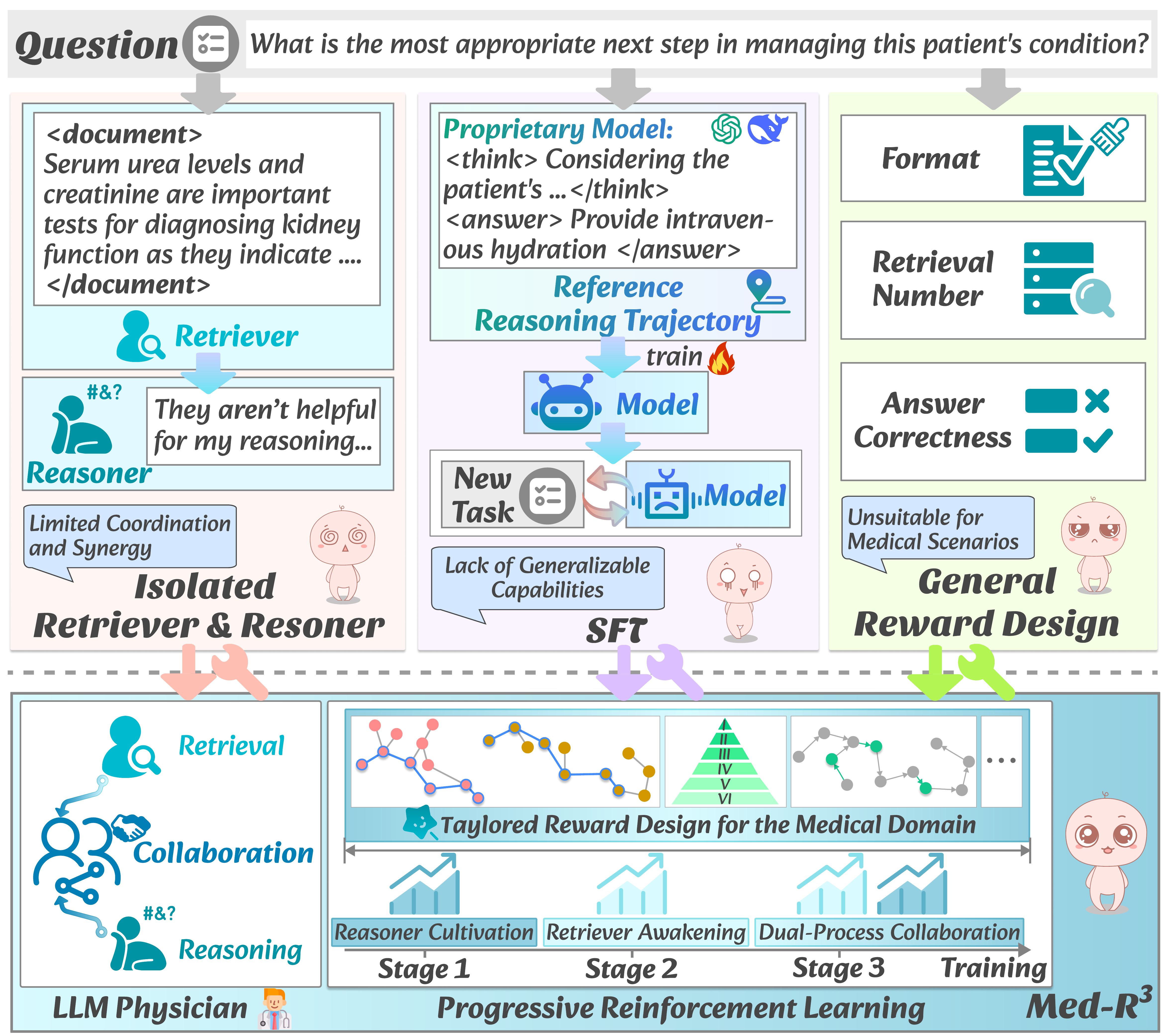}
    \caption{Comparison of \ours (\textit{bottom}) with existing methods (\textit{top}) for medical problem-solving.}
    \label{fig:intro}
    \vskip -0.1in
\end{figure}

\section{Introduction}
\label{sec:intro}

With the rapid development of artificial intelligence, large language models (LLMs) 
have shown remarkable potential 
in various fields~\cite{lewkowycz2022solving}. However, when applied to the medical domain, LLMs face unique challenges. 
Accurate diagnosis of medical conditions requires rigorous logical reasoning~\cite{lucas2024reasoning,savage2024diagnostic}, yet unlike domains such as mathematics or programming, where internal knowledge is often sufficient, the complexity and specificity of medical diagnosis necessitate the integration of external, up-to-date, and domain-specific knowledge~\cite{xiong2024benchmarking,lu2025med}. 
Therefore, during the process of solving medical problems, both \textit{retrieval} and \textit{reasoning} play crucial roles.

\begin{figure*}[t]
    \centering
    \includegraphics[width=1\textwidth]{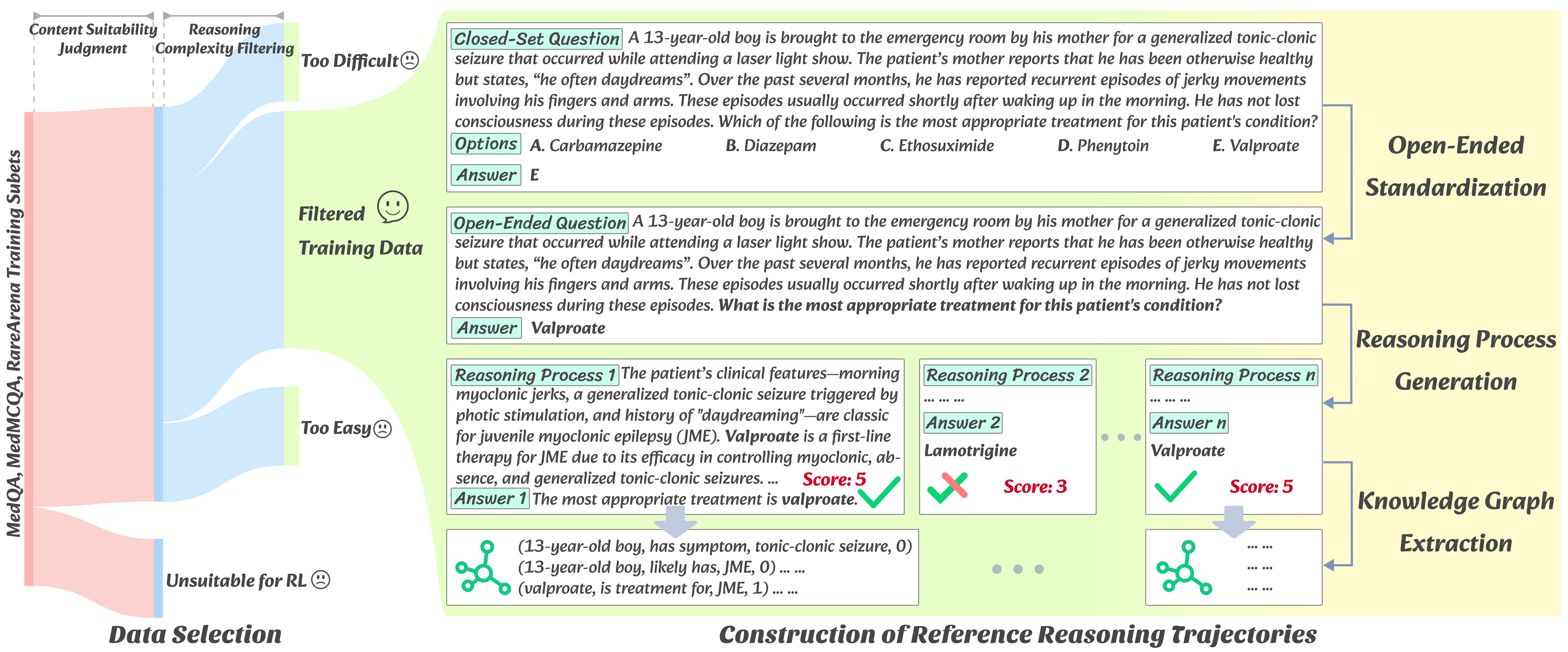}
    \caption{The pipeline of training data construction for Med-R$^3$.}
    \label{fig:data_construction}
\end{figure*}

Despite their potential, existing works face some limitations, as depicted in \Cref{fig:intro}: 
\textit{\textbf{(C1) Limited Coordination between Retrieval and Reasoning}}: 
While research has been devoted to independently 
improving the retrieval~\cite{jeong2024improving,xiong2024benchmarking,xiong2024improving} and reasoning~\cite{goh2024large,lucas2024reasoning} capabilities of models in the medical domain, 
they do not adequately align the retrieval strategy with requirements of the reasoning process, which is a critical gap that limits the system's end-to-end performance~\cite{yoran2023making,asai2024reliable}. 
\textit{\textbf{(C2) Lack of Generalization in SFT}}: 
Subsequent studies have investigated to achieve the joint optimization of these components during the supervised fine-tuning (SFT) stage~\cite{wang2024jmlr}. 
However, recent findings suggest that SFT inherently causes models to memorize task-specific shortcuts rather than learning generalizable reasoning in novel scenarios~\cite{lai2025med}. 
Conversely, reinforcement learning (RL) has shown high effectiveness at enhancing sophisticated reasoning capabilities in LLMs~\cite{jaech2024openai,guo2025deepseek,team2025kimi}. 
\textit{\textbf{(C3) Tailored Reward Design for Medical Scenarios}}: 
While recent works such as R1-Searcher~\cite{song2025r1} and ReSearch~\cite{chen2025learning} have explored training models for 
retrieval-augmented reasoning via reinforcement learning in general-domain tasks, their reward strategies are not well suited to the medical domain, where factors including the coverage of entities and relationships, as well as the credibility of retrieved documents, are of paramount importance in the reasoning process.

To address these challenges, we introduce \textbf{\textit{Med-R$^3$}}, a \underline{\textbf{\textit{Med}}}ical \underline{\textbf{\textit{R}}}etrieval-augmented \underline{\textbf{\textit{R}}}easoning framework driven by progressive \underline{\textbf{\textit{R}}}einforcement learning. 
For \textbf{\textit{C1}} and \textbf{\textit{C2}}, 
we first construct a medical training set of approximately 10K instances, each containing reference reasoning trajectories, and then 
we perform a progressive RL to co-optimize the model’s retrieval and reasoning capabilities in three distinct stages: 
\textbf{(1) Stage 1: Reasoner Cultivation.} 
We begin by developing the model’s logical reasoning abilities when addressing medical questions. 
\textbf{(2) Stage 2: Retriever Awakening.} 
Building upon the reasoning capabilities acquired in Stage 1, we adaptively optimize the retriever to better align with the retrieval features of the knowledge base and external information utilization during the model's reasoning process. 
\textbf{(3) Stage 3: Joint Optimization.} 
Finally, we refine the coordination between retrieval and reasoning in models to enhance their collaborative performance in medical scenarios. 
For \textbf{\textit{C3}}, we design specialized rewards tailored to the characteristics of medical reasoning at each training stage, encompassing aspects such as the coverage of entities and relations during the reasoning process, as well as the effectiveness of retrieval mechanisms that jointly consider the quality of medical evidence~\cite{sackett1996evidence} and the influence of retrieved documents within the overall reasoning trajectory. 
The main contributions are three-fold:

\begin{itemize}[leftmargin=*]
    \item \textit{\underline{Training Framework Advancement.}} 
    We propose Med-R$^3$, a 
    training framework to improve the retrieval-augmented reasoning performance 
    in the medical field 
    based on progressive reinforcement learning, 
    which jointly enhances the model's capability in retrieval and reasoning via a three-stage optimization strategy. 
    \item \textit{\underline{Reward Design Innovation.}} 
    Considering the unique characteristics of logical inference during medical problem-solving, we design reward metrics specifically tailored to the medical domain to supervise the reasoning process. 
    \item \textit{\underline{Performance and Effectiveness.}} 
    Extensive experiments indicate that models trained with \ours significantly improve medical performances. 
    Notably, Qwen3-8B + Med-R$^3$ surpasses the closed-sourced proprietary model GPT-4o-mini by 12.22\% at a comparable parameter scale, while Qwen2.5-14B integrated with \ours shows a more substantial gain of 16.31\%. 
\end{itemize}


\section{Med-R$^3$}
\label{sec:method}

\begin{figure*}[t]
    \centering
    \includegraphics[width=1\textwidth]{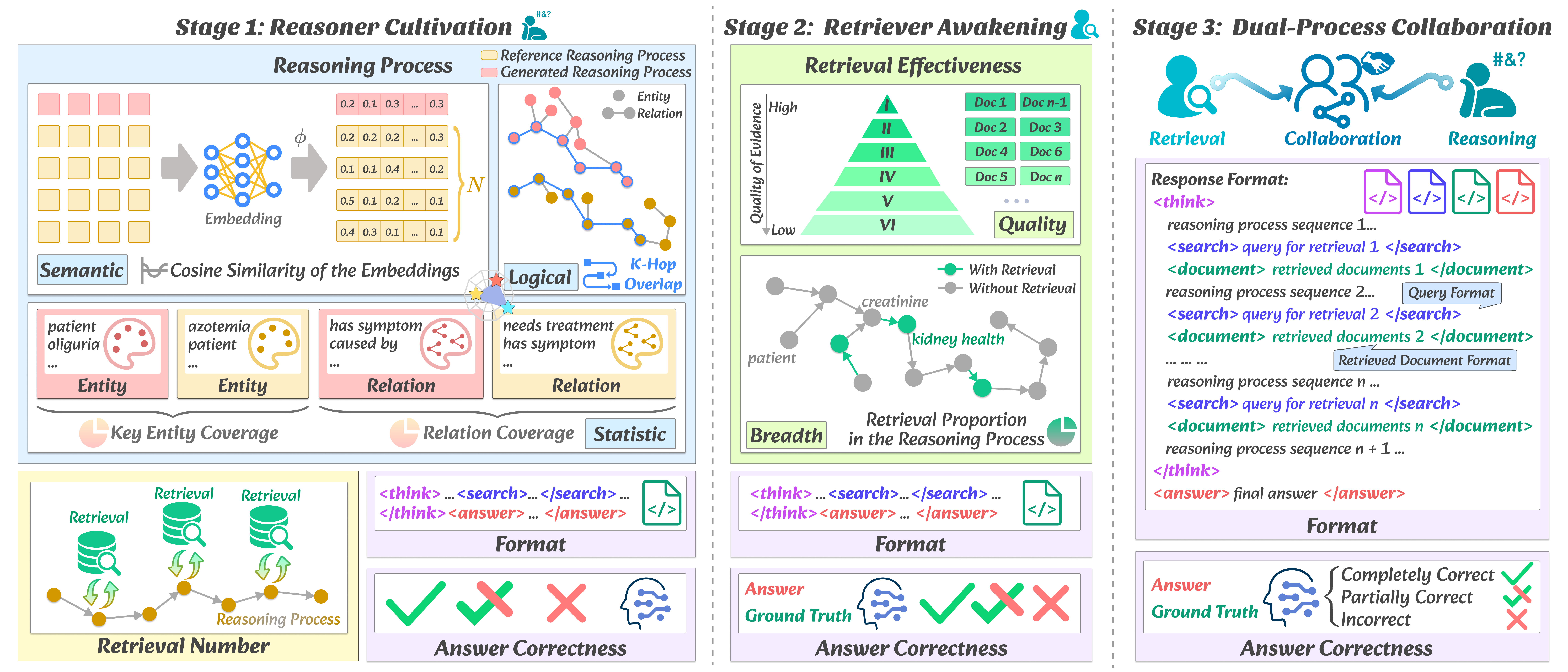}
    \caption{The progressive reinforcement learning pipeline of \textbf{\textit{Med-R$^3$}}. We designed tailored rewards for different training stages to facilitate the effective optimization of the model’s ability to 
    interpret and answer medical questions.}
    \label{fig:reinforcement_learning}
\end{figure*}

\subsection{Training Data Construction}
\label{subsec:training_data_construction}

\hypertarget{method}{Inspired} by the data selection and verifiability transformation process of HuatuoGPT-o1~\cite{chen2024huatuogpt}, we filtered medical questions based on their cognitive complexity and compatibility with the feature of reinforcement learning, and then constructed reference reasoning trajectories for verification, as illustrated in \Cref{fig:data_construction}, 
with details described in Appendix \ref{subsec:appendix_training_datasets}.

\subsubsection{Data Selection}

We utilize both closed-set exam questions and rare disease diagnostic datasets for our reinforcement learning phase, including the training sets from MedQA-USMLE, MedQA-MCMLE~\cite{jin2020disease}, MedMCQA~\cite{pal2022medmcqa}, and RareArena\footnote{For the RareArena dataset, we split the data into training and test sets with a ratio of 8:2 through random sampling.}~\cite{THUMedInfo_RareArena}. 
Data filtering and processing are conducted referring to the following criteria:

\begin{itemize}[leftmargin=*]
    \item \textit{Content Suitability Judgment}: 
    We employed the proprietary model, e.g., DeepSeek-V3~\cite{liu2024deepseek}, to exclude the medical questions that are unsuitable for RL or lacking unique answers (e.g., those requiring the identification of incorrect options).
    \item \textit{Reasoning Complexity Filtering}: 
    We then used 
    pass@n metrics to assess the question difficulty. 
    Specifically, we employed DeepSeek-R1~\cite{guo2025deepseek} to generate up to $n=16$ answers along with the corresponding reasoning process for each medical question. 
    DeepSeek-V3 was then utilized to evaluate both the intermediate reasoning steps and final answers, with scores ranging from 1 to 5. 
    We filtered out questions that were either too simple (achieving full correctness within a few rollouts, $n' \leq 3$) or excessively ambiguous (consistently receiving scores of 2 or lower across all 16 rollouts). 
    Finally, we obtained a training dataset comprising 2,140 questions from MedQA-USMLE, 1,204 from MedQA-MCMLE, 6,748 from MedMCQA, and 429 from the RareArena dataset. 
\end{itemize}

\subsubsection{Construction of Reference Reasoning Trajectories}

We performed data augmentation by rephrasing closed-set questions into open-ended formats, generating high-quality reasoning process, 
and further extracting structured reasoning paths that capture the 
pivotal logical steps leading to the final answer.

\begin{itemize}[leftmargin=*]
    \item \textit{Open-Ended Standardization}: We then reformatted the multiple-choice questions into open-ended formats using DeepSeek-V3~\cite{liu2024deepseek}, 
    transforming questions with predefined answer choices into tasks that require free-form, reasoning-based responses. 
    \item \textit{Reasoning Process Generation}: 
    Similar to the rollout process of data selection based on reasoning complexity, we prompted DeepSeek-R1 to generate all possible reasoning paths along with final answers for each medical question. 
    During the reasoning process, the model dynamically retrieves relevant external knowledge as needed and integrates the retrieved information to support continued inference. 
    Then we employed DeepSeek-V3 as an evaluator to assess the generated reasoning processes and answers on a scale from 1 to 5, retaining those with a score of 5. 
    Ultimately, for each medical question, we obtained an average of 5 reference reasoning processes. 
    \item \textit{Medical Knowledge Graph Extraction}: 
    Different from mathematical problems, which typically rely on rigid sequential logic, medical reasoning demands an understanding of rich, multi-relational structures that link symptoms, diagnoses, and therapeutic interventions. To better capture such a relational structure, we formalize reasoning processes as medical knowledge graph triplets, enabling a more precise comparison of coverage in both entities (e.g., disease, surgery, etc.) and relations. 
    Here we utilize DeepSeek-V3 to extract the medical knowledge graph from the natural language reasoning process: 
    \begin{gather}  
        \mathcal{G} = \{ (h, l, t, s) \mid h \in \mathcal{E}, l \in \mathcal{L}, t \in \mathcal{E}, s \in \mathbb{B} \}
        \label{eq:kg_retrieval_version}   
    \end{gather}
    where $\mathcal{E}$ denotes the set of all entities, $\mathcal{L}$ represents the set of all relations, and $(h, l, t)$ is a triple in which the head entity $h$ is connected to the tail entity $t$ via the relation $l$. 
    For each triplet, we add a binary attribute $s$, where $\mathbb{B} = \{0, 1\}$, specifying whether it was obtained through external 
    retrieval ($s = 1$) or internally generated via the model's reasoning capabilities ($s = 0$). 
\end{itemize}

\subsection{Progressive Reinforcement Learning}
\label{subsec:progressive_RL}

We employ a three-stage progressive pipeline to enhance the model’s capabilities in retrieval and reasoning for addressing problems in the medical scenario, 
as illustrated in \Cref{fig:reinforcement_learning}. 

\subsubsection{Stage 1: Cultivating the Reasoner}
\label{subsubsec:reasoner_training}

In this stage, we focus on developing the model’s ability to reason through medical problems. 
Unlike previous works~\cite{chen2024huatuogpt,liu2025beyond} that typically verify model performance based only on the correctness of the final answer, we incorporate a reasoning process evaluation mechanism into the reward function, aligning closely with the characteristics of clinical decision-making. 
Overall, the reward function design in Stage 1 
is \textit{the sum of normalized} 
components: 


    \textit{\textbf{Format.}} \quad 
    The model is required to produce its outputs according to a predefined output paradigm. Specifically: 
    \textit{\textbf{(1)}} All responses must strictly adhere to the format of \texttt{<think>...</think><answer>} \texttt{...</answer>}, and no duplicate tags are allowed within the response.  
    \textit{\textbf{(2)}} When retrieving the external knowledge is necessary, the corresponding query should be delineated using the \texttt{<search>...} \texttt{</search>} tag pair, 
    and then the retrieved documents from knowledge corpus are encapsulated within \texttt{<document>...</document>} tags. 
    Based on the above requirements, the reward is defined as: 
    
    \begin{gather}  
        \mathcal{R}_{{format}} = 
        \begin{cases}  
            1, \text{if the format is correct} \\  
            0, \text{if the format is incorrect} &
        \end{cases}
        \label{eq:stage1_format}    
    \end{gather}
    
    \textbf{\textit{Answer Correctness.}} \quad 
    In contrast to mathematical problems, medical questions often lack a strictly defined ground truth, as models may express the correct response with synonyms or paraphrases. As a result, exact string matching is \textit{not} an effective method for verifying correctness in medical tasks. 
    To address this, we employ a proprietary model (e.g., DeepSeek-V3) as the evaluator to score the generated answers. It assesses whether the model’s output semantically aligns with the reference ground truth. Answers are assigned a score of 2 for fully correct, 1 for partially correct (e.g., providing multiple answers, with at least one matching the correct answer), and 0 for incorrect responses: 
    
    \begin{gather}  
        \mathcal{R}_{answer} = 
        \begin{cases}  
            2, \text{if completely correct} \\  
            1, \text{if partially correct} \\
            0, \text{if incorrect} &
        \end{cases}
        \label{eq:stage1_answer}    
    \end{gather}
    
    \textit{\textbf{Reasoning Process.}} \quad 
    This aspect constitutes a core component in fostering the model’s reasoning capabilities during Stage 1. 
    Given the characteristics of medical reasoning, which requires a holistic understanding of interconnected entities and their relationships, e.g., those between symptoms, diagnoses, and therapeutic interventions~\cite{wu2023medical,gao2025leveraging}, we design the following reward for evaluation: 
    
    \begin{equation}
        \mathcal{R}_{reasoning} 
        = \tilde{R}_{semantic} + \tilde{R}_{statistic} + \tilde{R}_{logical}
        \label{eq:stage1_reasoning}    
    \end{equation} 
    where $\tilde{R}$ is the normalized score of $R$. 
    Given $N$ reference reasoning processes $\left\{r_{ref}^{(i)}\right\}_{i=1}^{N}$, 
    $R_{semantic}$, $R_{statistic}$ and $R_{logical}$ of the current generated response $r_{gen}$ are computed as follows.

\begin{algorithm}[ht]
\caption{Evaluation for the Logical Structure of the Reasoning Process (details for \Cref{eq:logical_score})}
\label{alg:logical_overlap}
\begin{flushleft}
\textbf{Input}: Model-generated reasoning paths $\mathcal{P}_{\text{gen}}$, reference reasoning trajectories $\mathcal{P}_{\text{ref}}$ \\
\textbf{Parameter}: Maximum path length $K$ for comparison \\
\textbf{Output}: Logical score of the reasoning process $ R_{\text{logical}} $ \\
\end{flushleft}
\begin{algorithmic}[1]
\STATE Initialize logical score: $ R_{\text{logical}} \gets 0 $
\STATE Compute maximize path length: 
$$ K \gets \min \{\max_{p \in \mathcal{P}_{\text{gen}}} |p|, \max_{p \in \mathcal{P}_{\text{ref}}} |p| \} 
$$

\FOR{each reference trajectory $ \mathcal{P}_{\text{ref}}^{(i)} \in \mathcal{P}_{\text{ref}} $}
    \STATE \mbox{Initialize the logical score for $\mathcal{P}_{\text{ref}}^{(i)}$: $ \text{Score}^{(i)} \gets 0 $} \\
    \FOR{$ j = 1, 2, ..., K $}
        \STATE Filter model paths of length $ j $: 
        $$ \mathcal{P}_{\text{gen}}^{(j)} \gets \{ p \in \mathcal{P}_{\text{gen}} \mid |p| = j \} 
        $$ 
        $$ \mathcal{P}_{\text{ref}}^{(ij)} \gets \{ p \in \mathcal{P}_{\text{ref}}^{(i)} \mid |p| = j \} 
        $$ 
        \IF{$ \mathcal{P}_{\text{gen}}^{(j)} \neq \emptyset $}
            \STATE Compute Jaccard similarity for the $ j $-hop paths and update weighted score:
            $$
            \text{Score}^{(i)} \gets \text{Score}^{(i)} + j \cdot  \frac{|\mathcal{P}_{\text{ref}}^{(ij)} \cap \mathcal{P}_{\text{gen}}^{(j)}|}{|\mathcal{P}_{\text{ref}}^{(ij)} \cup \mathcal{P}_{\text{gen}}^{(j)}|}
            $$
        \ENDIF
    \ENDFOR
    \STATE Normalize the logical score for $\mathcal{P}_{\text{ref}}^{(i)}$: $$ \text{Score}^{(i)} = \text{Score}^{(i)} \cdot \frac{1}{\sum_{j=1}^{K} 
            j} = \text{Score}^{(i)} \cdot \frac{2}{K(K+1)} $$
    \STATE Update $ R_{\text{logical}} \gets \max \{ R_{\text{logical}}, \text{Score}^{(i)} \} $
\ENDFOR
\RETURN $ R_{\text{logical}} $
\end{algorithmic}
\end{algorithm}

    \begin{itemize} 
        \item $R_{semantic}$: 
        This aspect measures the semantic alignment between the reasoning process generated by the model and those previously constructed by the proprietary model for reference, which is quantified by the cosine similarity~\cite{salton1975vector}. $\phi (r)$ denotes the embedding of the reasoning process utilized the embedding model~\cite{chen2024bge}.
        
        \begin{equation}
            R_{semantic} = \underset{i \in [1, N]}{\max} \, cos \Big( \phi \big( r_{\text{ref}}^{(i)} \big), \, \phi \big( r_{\text{gen}} \big) \Big)
        \label{eq:semantic_score}
        \end{equation} 
        \item $R_{statistic}$: 
        Medical reasoning typically emphasizes the completeness of inferred entities and their interrelations. 
        To capture this aspect, we introduce evaluation metrics based on the coverage of entities and relations. 
        Following the procedure used during training data construction, we extracted knowledge graphs from the natural language reasoning processes generated by the model, resulting in triplets in the form of \Cref{eq:kg_retrieval_version}.  
        We then compute the Jaccard similarity~\cite{jaccard1912distribution} between the entity and relation sets generated by the model during training and those present in the reference reasoning processes.
        \begin{equation}
            R_{statistic} = \underset{i \in [1, N]}{\max} 
            \left( \underbrace{\frac{|\mathcal{E}_{\text{ref}}^{(i)} \cap \mathcal{E}_{\text{gen}}|}{|\mathcal{E}_{\text{ref}}^{(i)} \cup \mathcal{E}_{\text{gen}}|}}_{\text{Key Entity Coverage}} + \underbrace{\frac{|\mathcal{L}_{\text{ref}}^{(i)} \cap \mathcal{L}_{\text{gen}}|}{|\mathcal{L}_{\text{ref}}^{(i)} \cup \mathcal{L}_{\text{gen}}|}}_{\text{Relation Coverage}} \right)
        \label{eq:statistic_score}
        \end{equation}
        
        \item $R_{logical}$: 
        To evaluate the multi-step logical structure of reasoning, 
        we also assess the logical alignment by computing the Jaccard similarity between the $j$-hop ($j=1,2,..., K$) reasoning paths generated by the model during training, denoted as $\mathcal{P}_{\text{gen}}^{(j)}$, and the $i$-th reference diagnostic reasoning trajectories $\mathcal{P}_{\text{ref}}^{(ij)}$, with the algorithm implementation provided in \Cref{alg:logical_overlap}. 
        Here $K$ is the minimum path length between the longest reasoning chains extracted from the model-generated and reference medical knowledge graphs.
        \begin{equation}
            R_{logical} = \underset{i \in [1, N]}{\max} \, \frac{2}{K(K+1)} \sum_{j=1}^{K} 
            j \cdot \frac{|\mathcal{P}_{\text{ref}}^{(ij)} \cap \mathcal{P}_{\text{gen}}^{(j)}|}{|\mathcal{P}_{\text{ref}}^{(ij)} \cup \mathcal{P}_{\text{gen}}^{(j)}|}
        \label{eq:logical_score}
        \end{equation}                  
    \end{itemize}
    
    \textit{\textbf{Retrieval Number.}} \quad 
    Another objective of this stage is to encourage the model to conduct external knowledge searches to assist its reasoning process, thereby laying the foundation for improving retrieval capabilities in the subsequent stage. 
    To this end, 
    we established a reward mechanism based on the number of search operations performed: 
    \begin{gather}  
        \mathcal{R}_{retrieval\_num} = 
        \begin{cases} 
            1, \text{if} \quad n \ge \delta \\
            0, \text{if} \quad n < \delta &
        \end{cases}
        \label{eq:stage1_retrieval} 
    \end{gather}
    where $n$ indicates the count of retrieval invocations, and the minimum allowable number of searches is set to $\delta = 3$, which is based on the actual statistical results of the average retrieval count when DeepSeek-R1 generated reference reasoning trajectories during the data construction phase in \Cref{subsec:training_data_construction}.

\subsubsection{Stage 2: Awakening the Retriever}
\label{subsubsec:retriever_training}

Following the initial training stage, the model has acquired a foundational paradigm for medical reasoning. 
In this stage, we shift our focus to enhancing the model’s ability to retrieve external knowledge.   
Specifically, we aim to improve 
the generation of semantically accurate and retrieval-efficient query terms. 
In general, the design of the reward function in Stage 2 
is \textit{the sum of normalized} 
components: 
\textbf{\textit{format}}, \textbf{\textit{answer correctness}}, and \textbf{\textit{retrieval effectiveness}}, 
with the first two already formally defined in \Cref{eq:stage1_format} and \Cref{eq:stage1_answer}.

\textit{\textbf{Retrieval Effectiveness.}} \quad 
It encompasses two 
normalized reward components: 
\textbf{\textit{(1)}} the authority of the retrieved medical documents, and \textbf{\textit{(2)}} the extent to which the retrieved content is utilized throughout the entire reasoning process: 

\begin{equation}
    \mathcal{R}_{retrieval} 
    = \tilde{R}_{quality} + \tilde{R}_{breadth}
    \label{eq:stage2_reasoning}    
\end{equation}

\begin{itemize} 
    \item $R_{quality}$: 
    According to the principles of Evidence-Based Medicine (EBM)~\cite{sackett1996evidence}, not all evidence plays equal roles in clinical decision-making, where the hierarchy of evidence is critical, as depicted in \Cref{fig:hierarchy_of_evidence} (details are provided in Appendix \ref{subsubsec:EBM_appendix}). 
    Each retrieved document is assigned an integer evidence level $e$ by the proprietary model (e.g., DeepSeek-V3), 
    where $e \in \left \{ x \in \mathbb{Z} \mid 1 \leq x \leq 6 \right \}$, with 1 indicating the highest level of credibility. 
    Here $D$ denotes the number of documents retrieved by the model during a single rollout: 
    \begin{gather}  
        R_{quality} = \frac{1}{D} \sum_{j=1}^{D} (6 - \left( e_j - 1\right ))
        \label{eq:authority}    
    \end{gather}
    
    \item $R_{breadth}$: 
    We also follow the procedure during data construction, where we extract knowledge graph triplets in the form of \Cref{eq:kg_retrieval_version} from the reasoning processes generated by the model. 
    Therefore, the proportion of retrieved triplets in the reasoning process is: 
    
    \begin{gather}
        R_{breadth} = \frac{1}{|\mathcal{G}|} \sum_{j=1}^{|\mathcal{G}|} s_j = \frac{\sum_{(h,l,t,s) \in \mathcal{G}} \mathbb{I}(s=1)}{|\mathcal{G}|}
        \label{eq:breadth}
    \end{gather}
\end{itemize}

\subsubsection{Stage 3: Orchestrating the Dual-Process Collaboration}
\label{subsubsec:retriever_reasoner_training}

Having separately enhanced the model’s reasoning and retrieval capabilities in the earlier stages, Stage 3 concentrates on strengthening the coordination between reasoning and retrieval. 
The reward function at this stage is \textit{the sum of normalized} \textbf{\textit{format}} and \textbf{\textit{answer correctness}}, which directly prioritizes end-to-end (E2E) performance of the ultimate objective, as defined in \Cref{eq:stage1_format} and \Cref{eq:stage1_answer}.

\section{Experiments}
\label{sec:experiments}

\subsection{Experimental Setup}
\label{subsec:experimental_setup}

\textbf{Datasets.} \quad 
Datasets for \textit{training} are detailed in \hyperlink{method}{the previous section}. 
For \textit{evaluation}, we have selected 8 medical datasets including the MedQA-USMLE, MedQA-MCMLE (hereafter, abbreviated as MedQA-US and MedQA-MC, respectively)~\cite{jin2020disease}, MedMCQA~\cite{pal2022medmcqa}, 
RareArena\footnote{We use the combined dev and test subsets for MedQA, the dev subset for MedMCQA, and our pre-split test sets for RareArena.}~\cite{THUMedInfo_RareArena}, 
MMLU-Med~\cite{hendryckstest2021}, NEJMQA~\cite{katz2024gpt}, MedXpertQA~\cite{zuo2025medxpertqa} and HealthBench~\cite{arora2025healthbench}, 
covering both standard and real-world clinical scenarios. 
We utilize \textit{LLM-as-Judge}~\citep{zheng2023judging,gu2024survey} 
based on the proprietary model DeepSeek-V3~\cite{liu2024deepseek} 
to verify the correctness of the responses, and then calculate the accuracy scores as the evaluation metric. 

\begin{figure}[t]
    \centering
    \includegraphics[width=1\linewidth]{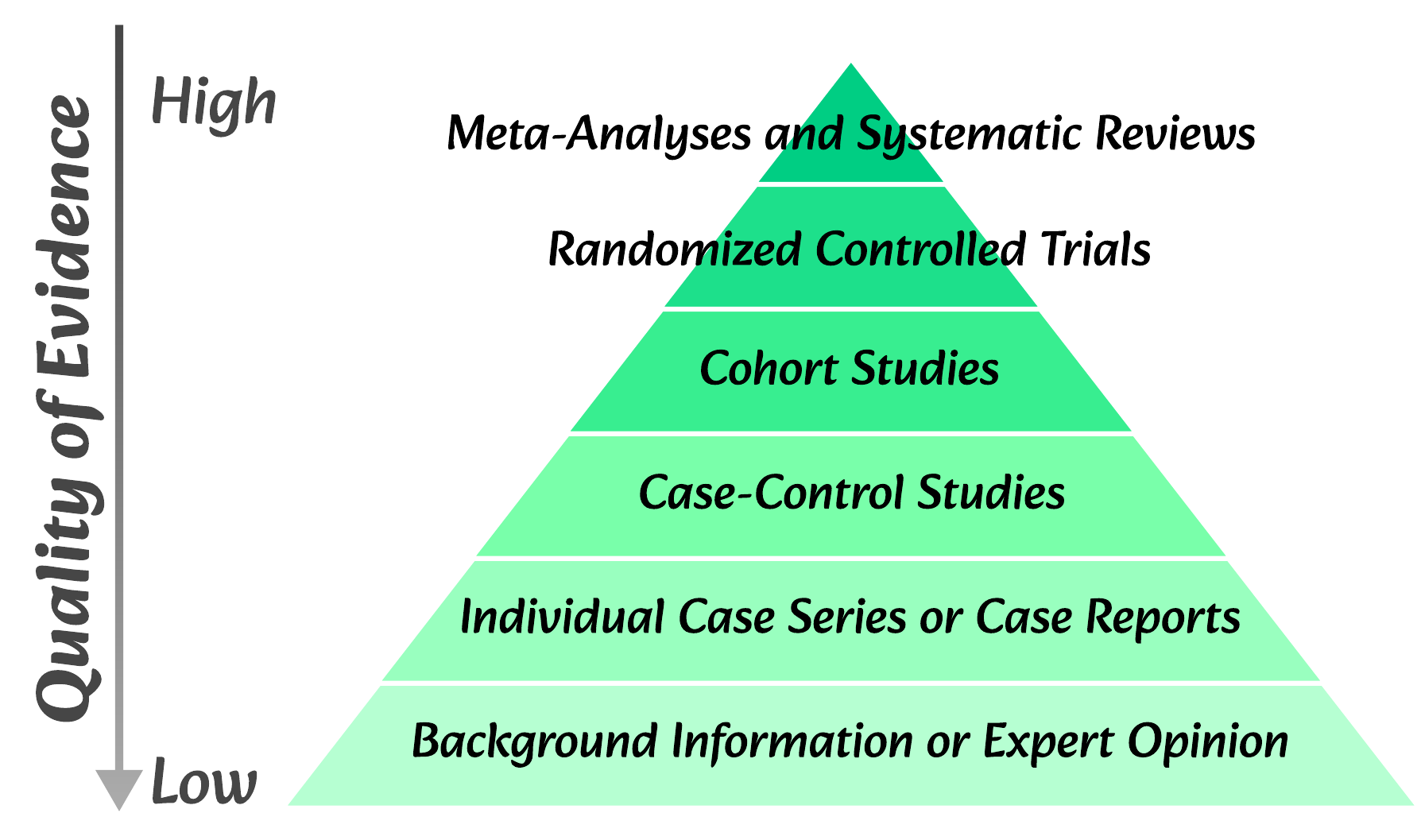}
    \caption{Illustration of \textit{Hierarchy of Evidence} in EBM.}
    \label{fig:hierarchy_of_evidence}
\end{figure}

\begin{table}[t]
    \centering
    \caption{\label{tab:medical_knowledge_corpus}
            Overall statistics of medical knowledge resources.}
    \resizebox{\linewidth}{!}{
        \begin{tabular}{cccc}
            \toprule
            Source Type & Data Resource & \#Volume & \#Avg. Token \\ 
            \midrule
            Papers & PubMed Central~\cite{roberts2001pubmed} &  600,000 & 3,820 \\ 
            Entries & Wikipedia~\cite{wikidump} & 470,000 & 1,387 \\
            Books & NCBI Bookshelf~\cite{hoeppner2012ncbi} & 10,000 & 4,083 \\
            Guidelines & Guidelines from \citet{chen2023meditron} & 10,000 & 1,100 \\
            \bottomrule
        \end{tabular}
    }
\end{table}

\begin{table*}[tb]
    \centering
    \caption{\label{tab:main_results} 
    Comparison of \ours with baselines. ``$*$'' denotes the re-implementation with the same amount of our constructed data for a fair comparison. The best and second best scores of each model are in \textbf{bold} and \underline{underlined}. 
} 
    \setlength{\tabcolsep}{1.4pt}
    \resizebox{\textwidth}{!}{
        \begin{tabular}{c|ccccccccccc}
            \toprule
                \multirow{2}{*}{Model} & \multirow{2}{*}{Method} & \multicolumn{1}{|c}{MedQA-US} & MedQA-MC & MedMCQA & RareArena-RDC & RareArena-RDS & \multicolumn{1}{|c}{MMLU-Med} & NEJMQA & MedXpertQA & \multicolumn{1}{c|}{HealthBench} & \multirow{2}{*}{\textbf{Avg.}} \\
            \cline{3-11}
             & & \multicolumn{5}{|c}{\textit{In-Domain (ID)}} &  \multicolumn{4}{|c|}{\textit{Out-of-Domain (OOD)}} &  \\
            \midrule
            \multicolumn{12}{c}{\cellcolor[HTML]{EFEFEF}\textbf{\textit{Close-Sourced Models}}} \\
            \midrule
            GPT-4o-mini & -- & \multicolumn{1}{|c}{74.45} & 69.25 & 70.52 & 52.03 & 43.24 & \multicolumn{1}{|c}{80.96} & 57.87 & 21.58 & 28.50 & \multicolumn{1}{|c}{55.38} \\
            \midrule
            \multicolumn{12}{c}{\cellcolor[HTML]{EFEFEF}\textbf{\textit{Open-Sourced Medical-Specific Models}}} \\
            \midrule
            MEDITRON-7B & -- & \multicolumn{1}{|c}{48.67} & 44.28 & 46.78 & 37.97 & 21.60 & \multicolumn{1}{|c}{50.12} & 33.40 & 16.55 & 15.50 & \multicolumn{1}{|c}{34.98} \\
            UltraMedical3-8B & -- & \multicolumn{1}{|c}{61.57} & 52.42 & 61.82 & 40.76 & 28.54 & \multicolumn{1}{|c}{72.52} & 45.31 & 10.82 & 21.30 & \multicolumn{1}{|c}{43.90} \\
            UltraMedical3.1-8B & -- & \multicolumn{1}{|c}{66.86} & 58.45 & 65.73 & 43.83 & 32.39 & \multicolumn{1}{|c}{75.86} & 50.66 & 12.08 & 24.10 & \multicolumn{1}{|c}{47.77} \\
            MEDITRON-70B & -- & \multicolumn{1}{|c}{60.60} & 55.64 & 56.48 & 75.16 & 48.85 & \multicolumn{1}{|c}{70.53} & 65.33 & 18.72 & 31.20 & \multicolumn{1}{|c}{53.61} \\
            \midrule
            \multicolumn{12}{c}{\cellcolor[HTML]{EFEFEF}\textbf{\textit{Open-Sourced Medical Reasoning Models}}} \\
            \midrule
            HuatuoGPT-o1-8B & -- & \multicolumn{1}{|c}{66.97} & 66.15 & 72.45 & 49.76 & 41.59 & \multicolumn{1}{|c}{74.52} & 51.60 & 14.34 & 31.80 & \multicolumn{1}{|c}{52.13} \\
            MedS$^3$-8B & -- & \multicolumn{1}{|c}{73.51} & 69.63 & 65.47 & 46.85 & 37.72 & \multicolumn{1}{|c}{78.75} & 55.09 & 12.50 & 33.20 & \multicolumn{1}{|c}{52.52} \\
            AlphaMed-8B & -- & \multicolumn{1}{|c}{64.06} & 64.98 & 66.43 & 43.55 & 38.14 & \multicolumn{1}{|c}{71.44} & 51.48 & 22.01 & 32.60 & \multicolumn{1}{|c}{50.52} \\
            MedResearcher-R1-32B & -- & \multicolumn{1}{|c}{76.42} & 78.15 & 73.58 & 62.21 & 50.93 & \multicolumn{1}{|c}{83.14} & 58.47 & 15.82 & 35.70 & \multicolumn{1}{|c}{59.38} \\
            \midrule
            \multicolumn{12}{c}{\cellcolor[HTML]{EFEFEF}\textbf{\textit{Open-Sourced Base / Instruct Models}}} \\
             \midrule
             \multirow{5}{*}{\makecell{LLaMA3.1-8B\\-Instruct}} & Naive Response & \multicolumn{1}{|c}{31.16} & 41.45 & 30.02 & 41.85 & 22.16 & \multicolumn{1}{|c}{37.12} & 50.41 & 14.90 & 16.50 & \multicolumn{1}{|c}{31.73} \\
             & SFT & \multicolumn{1}{|c}{61.39} & 62.10 & 63.27 & 44.83 & 35.20 & \multicolumn{1}{|c}{63.08} & 49.64 & 12.72 & 31.80 & \multicolumn{1}{|c}{47.11} \\
             & R1-Searcher$*$ & \multicolumn{1}{|c}{60.28} & 60.92 & 63.54 & 42.87 & \underline{38.65} & \multicolumn{1}{|c}{70.19} & \underline{53.81} & \underline{15.73} & 34.60 & \multicolumn{1}{|c}{48.95} \\
             & ReSearch$*$ & \multicolumn{1}{|c}{\underline{62.76}} & \underline{66.03} & \underline{66.25} & \underline{46.35} & 38.44 & \multicolumn{1}{|c}{\underline{71.27}} & 53.26 & 14.65 & \underline{38.70} & \multicolumn{1}{|c}{\underline{50.86}} \\
             & \textbf{\ours (ours)} & \multicolumn{1}{|c}{\textbf{75.91}} & \textbf{75.95} & \textbf{75.89} & \textbf{57.34} & \textbf{47.16} & \multicolumn{1}{|c}{\textbf{79.07}}  & \textbf{60.60} & \textbf{16.48} & \textbf{42.80} & \multicolumn{1}{|c}{\textbf{59.02}} \\
             \midrule
             \multirow{5}{*}{Qwen2.5-7B} & Naive Response & \multicolumn{1}{|c}{22.58} & 39.14 & 28.77 & 32.17 & 23.10 & \multicolumn{1}{|c}{44.45} & 41.48 & 11.59 & 14.60 & \multicolumn{1}{|c}{28.65} \\
             & SFT & \multicolumn{1}{|c}{52.56} & 50.04 & 57.90 & 53.45 & 34.67 & \multicolumn{1}{|c}{56.94}  & 48.23 & 11.28 & 29.70 & \multicolumn{1}{|c}{43.86} \\
             & R1-Searcher$*$ & \multicolumn{1}{|c}{56.78} & 49.70 & 58.35 & 53.69 & 33.27 & \multicolumn{1}{|c}{66.81} & \underline{52.98} & 12.55 & 32.90 & \multicolumn{1}{|c}{46.34} \\
             & ReSearch$*$ & \multicolumn{1}{|c}{\underline{62.47}} & \underline{60.24} & \underline{63.11} & \underline{55.95} & \underline{34.68} & \multicolumn{1}{|c}{\underline{70.29}} & 52.30 & \underline{12.67} & \underline{38.10} & \multicolumn{1}{|c}{\underline{49.98}} \\
             & \textbf{\ours (ours)} & \multicolumn{1}{|c}{\textbf{68.64}} & \textbf{67.53} & \textbf{68.97} & \textbf{63.02} & \textbf{45.76} & \multicolumn{1}{|c}{\textbf{75.81}}  & \textbf{58.54} & \textbf{14.98} & \textbf{41.00} & \multicolumn{1}{|c}{\textbf{56.03}} \\
             \midrule
             \multirow{5}{*}{Qwen3-8B} & Naive Response & \multicolumn{1}{|c}{38.43} & 45.62 & 35.25 & 36.54 & 21.81 & \multicolumn{1}{|c}{55.47} & 48.23 & 13.56 & 18.50 & \multicolumn{1}{|c}{34.82} \\
             & SFT & \multicolumn{1}{|c}{61.54} & 59.83 & 62.45 & \underline{64.21} & 38.62 & \multicolumn{1}{|c}{66.54} & 50.36 & \underline{14.82} & 33.70 & \multicolumn{1}{|c}{50.23} \\
             & R1-Searcher$*$ & \multicolumn{1}{|c}{59.82} & 61.24 & 64.15 & 62.53 & 40.42 & \multicolumn{1}{|c}{71.21} & \underline{56.84} & 14.23 & 36.40 & \multicolumn{1}{|c}{51.87} \\
             & ReSearch$*$ & \multicolumn{1}{|c}{\underline{64.63}} & \underline{62.54} & \underline{65.32} & 63.81 & \underline{41.63} & \multicolumn{1}{|c}{\underline{73.42}} & 54.93 & 14.54 & \underline{40.30} & \multicolumn{1}{|c}{\underline{53.46}} \\
             & \textbf{\ours (ours)} & \multicolumn{1}{|c}{\textbf{76.84}} & \textbf{74.92} & \textbf{76.53} & \textbf{75.62} & \textbf{53.24} & \multicolumn{1}{|c}{\textbf{82.83}} & \textbf{59.42} & \textbf{15.14} & \textbf{44.80} & \multicolumn{1}{|c}{\textbf{62.15}} \\
             \midrule
             \multirow{5}{*}{Qwen2.5-14B} & Naive Response & \multicolumn{1}{|c}{50.01} & 50.70 & 42.85 & 43.17 & 26.58 & \multicolumn{1}{|c}{71.60} & 45.63 & 11.06 & 22.40 & \multicolumn{1}{|c}{40.44} \\
             & SFT & \multicolumn{1}{|c}{68.85} & 70.22 & 70.15 & 75.27 & 52.82 & \multicolumn{1}{|c}{75.81}  & 47.08 & 11.54 & 37.70 & \multicolumn{1}{|c}{56.61} \\
             & R1-Searcher$*$ & \multicolumn{1}{|c}{69.20} & 71.75 & 68.45 & \underline{76.32} & \underline{54.05} & \multicolumn{1}{|c}{77.69} & \underline{52.08} & 12.65 & \underline{43.90} & \multicolumn{1}{|c}{58.45} \\
             & ReSearch$*$ & \multicolumn{1}{|c}{\underline{69.52}} & \underline{74.05} & \underline{72.20} & 75.67 & 53.54 & \multicolumn{1}{|c}{\underline{80.25}} & 50.65 & \underline{13.10} & 41.30 & \multicolumn{1}{|c}{\underline{58.92}} \\
             & \textbf{\ours (ours)} & \multicolumn{1}{|c}{\textbf{78.01}} & \textbf{80.59} & \textbf{75.42} & \textbf{77.94} & \textbf{58.15} & \multicolumn{1}{|c}{\textbf{85.33}} & \textbf{62.40} & \textbf{15.69} & \textbf{46.20} & \multicolumn{1}{|c}{\textbf{64.41}} \\
            \bottomrule
        \end{tabular}
    }
\end{table*}

\textbf{Medical Knowledge Corpus.} \quad 
We establish a comprehensive medical knowledge base to support information retrieval, curated from multiple data reliable sources~\cite{roberts2001pubmed,hoeppner2012ncbi,chen2023meditron,wikidump,lu2025med}. 
It comprises four representative types of resources: \textit{academic papers}, \textit{entries}, \textit{books}, and \textit{guidelines}, offering both depth and breadth of information. 
We employ the hybrid retrieval strategy, including BGE-Large-EN-v1.5~\cite{bge_embedding} for dense and SPLADE-v3~\cite{lassance2024splade} 
for sparse retrieval. 
During the rollout phase in both training and evaluation, we retrieve the top-5 related documents for each query. 
Data statistics are outlined in \Cref{tab:medical_knowledge_corpus}, and details can be found in Appendix \ref{subsec:appendix_knowledge_corpus}.

\textbf{Models and Implementation.} 
We use LLaMA3.1-8B-Instruct~\cite{dubey2024llama}, Qwen3-8B~\cite{yang2025qwen3}, 
Qwen2.5-7B and Qwen2.5-14B~\cite{yang2024qwen2} 
as the backbone models for our training. 
The \textit{reinforcement learning (RL)} framework is built on verl~\cite{sheng2025hybridflow} with Group Relative Policy Optimization (GRPO)~\cite{shao2024deepseekmath} as the learning algorithm. 
The training dataset obtained from the data construction phase contains 10,521 samples. Each sample undergoes 16 rollouts during training, with a training batch size of 256 and a rollout batch size of 64. The total number of training epochs is set to 3, where each training stage corresponds to 1 epoch comprising 41 training steps. 
The learning rate is 1e-6. 
Notably, in our training setup, external documents retrieved by the model are concatenated into the reasoning process, which are not generated by the training policy. To prevent these retrieved segments from influencing gradient updates, we apply masking during loss calculation, where we mask out all content enclosed within \texttt{<document>...</document>} tags. 
When conducting \textit{supervised fine-tuning (SFT)} for method comparison, we utilized a learning rate scheduler featuring linear warm-up and cosine decay, with a learning rate peaking at 2e\mbox{-}5, alongside a warmup ratio of 0.03 and a weight decay of 0.0. 
Our code is available in the repository \href{https://anonymous.4open.science/r/Med-R3}{https://anonymous.4open.science/r/Med-R3}.

\textbf{Baselines.} \quad 
We compare our \ours method against the following baselines: 
(1) We employ GPT-4o-mini~\cite{hurst2024gpt} as the \textbf{\textit{Close-Sourced Models}} competitors. 
(2) \textbf{\textit{Open-Sourced Medical-Specific Models}} include MEDITRON-7B, MEDITRON-70B~\cite{chen2023meditron}, UltraMedical3-8B, and UltraMedical3.1-8B~\cite{zhang2024ultramedical}. 
(3) \textbf{\textit{Open-Sourced Medical Reasoning Models}} comprise HuatuoGPT-o1-8B~\cite{chen2024huatuogpt},  MedS$^3$-8B~\cite{jiang2025meds} and AlphaMed-8B~\cite{alhamed2025}, 
all of which employ LLaMA3.1-8B-Instruct~\cite{dubey2024llama} as their backbone, together with MedResearcher-R1-32B~\cite{yu2025medresearcher}, whose backbone is Qwen2.5-32B. 
(4) The simplest baseline is \textbf{\textit{Naive Response}}, where the model generates responses directly without external knowledge retrieval or training. 
(5) We also compare with the training strategy in which we perform \textbf{\textit{Supervised Fine-Tuning (SFT)}} using our constructed reasoning trajectories produced by proprietary models. 
(6) \textbf{\textit{General Retrieval-Augmented Reasoning RL}} for comparison includes R1-Searcher~\cite{song2025r1} and ReSearch~\cite{chen2025learning}. 
More details of these baselines can be found in Appendix \ref{subsec:appendix_baseline}.

\subsection{Main Results}

The main results of baselines and \ours are demonstrated in \Cref{tab:main_results}, and we summarize the observations below. 

\textbf{\ours is effective across different models.} 
Experimental 
results in \Cref{tab:main_results} show that \ours consistently outperforms other baseline methods on both base and instruction-tuned 
models across different scales in terms of medical problem-solving. 
Compared to \textit{naive response}, \ours enhances average downstream task performances by 78.13\%. 
As opposed to \textit{supervised fine-tuning (SFT)}, our method yields a 22.14\% gain, 
and it exceeds the average performance of \textit{general retrieval-augmented reasoning RL} approaches by 15.37\%. 
Furthermore, models enhanced with \ours demonstrate the potential to outperform \textit{close-sourced models} in medical scenarios. Specifically, Qwen3-8B + \ours achieves an average improvement of 12.22\% over GPT-4o-mini at a comparable parameter scale, while Qwen2.5-14B + \ours shows a more substantial gain of 16.31\%. 
Remarkably, even when compared to larger \textit{open-sourced medical-specific models} such as MEDITRON-70B~\cite{chen2023meditron}, our approach achieves over 12.67\% higher performance on average with much smaller parameter scale (7B–14B).

\begin{figure}[tb]
    \centering
    \includegraphics[width=1\linewidth]{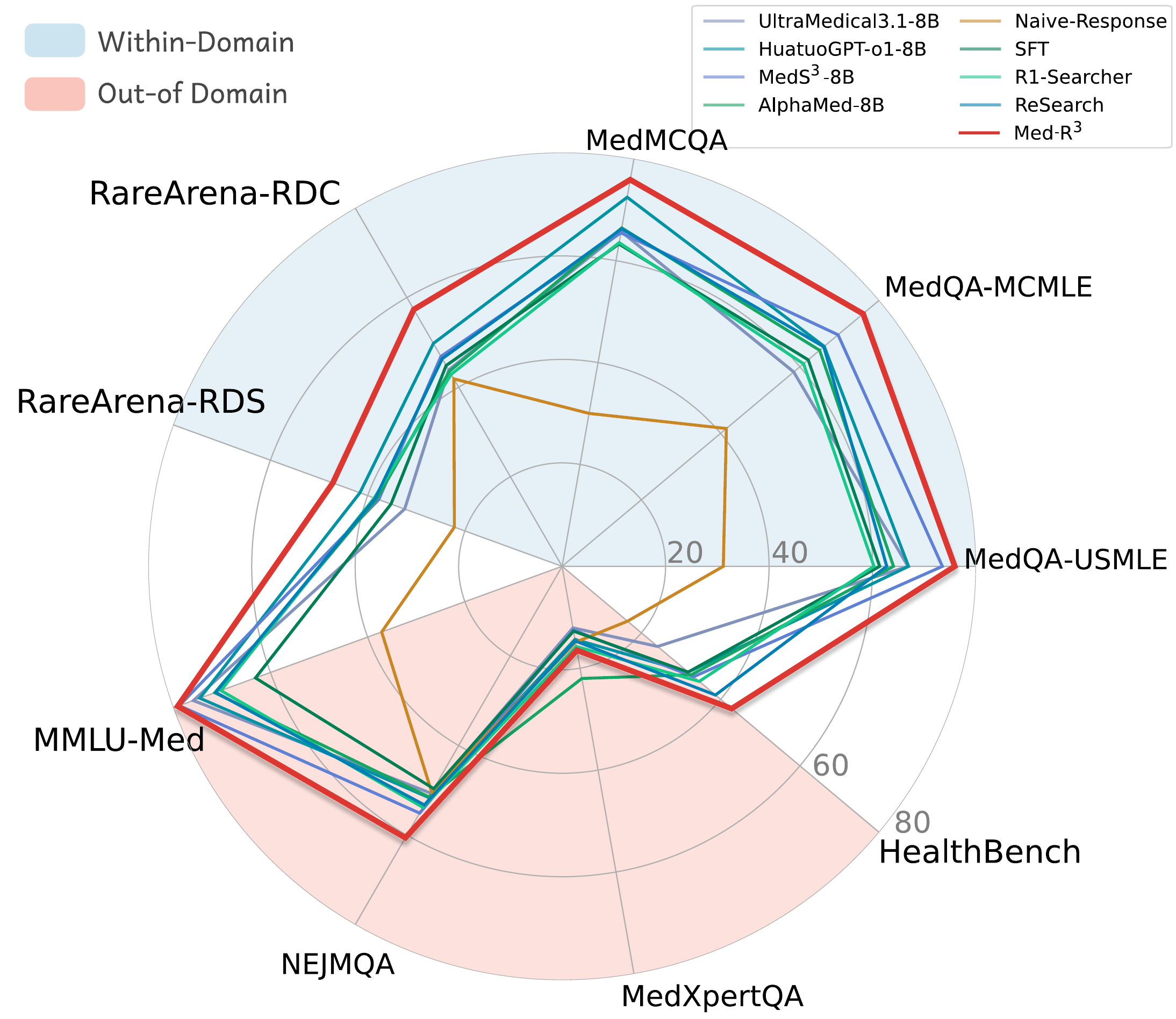}
    \caption{Comparison of \ours with baselines that utilize the \textbf{\textit{LLaMA3.1-8B-Instruct} as the same backbone model.}}
    \label{fig:radarchart_llama3}
    \vspace{-0.3cm}
\end{figure}

\textbf{Retrieval-augmented-reasoning boosts medical performa- nces.} 
The backbone model for most of \textit{open-sourced medical reasoning models} considered in this study is LLaMA3.1-8B-Instruct~\cite{dubey2024llama}. 
One of the advantages of \ours lies in its ability to dynamically retrieve and incorporate external medical knowledge during the reasoning process. 
Experimental results demonstrate that \ours achieves an average performance improvement of over 14.11\% across medical tasks compared to these competitors. 
In addition, we use the radar chart to visualize the performance of the methods that share the same backbone model of LLaMA3.1-8B-Instruct, as depicted in \Cref{fig:radarchart_llama3}. 
This finding further underscores the importance of integrating external knowledge retrieval during reasoning, particularly in the medical domain, where knowledge is highly specialized, rapidly evolving, and broad in scope.

\section{Ablations and Analysis}
\label{sec:ablation_study}

\begin{table}[tb] 
    \centering
    \caption{\label{tab:ablation}
    Analysis of the training stages and sequential order using \textbf{\textit{Qwen2.5-7B}}. ``Order'': the sequence of Stage 1, 2, and 3. 
    ``\underline{1 $\&$ 2 $\&$ 3}'' denotes a training setting in which the reward functions from all three stages are applied simultaneously. 
    We compute the \textit{average} performance of the evaluated models across in-domain and out-of-domain benchmarks.
    The best and second best scores are in \textbf{bold} and \underline{underlined}.
    }
	\resizebox{1\linewidth}{!}{
		\begin{tabular}{c|c|c|c}
			\toprule
			Order & In-Domain & Out-of-Domain & Avg. \\ 
            \midrule
            \multicolumn{4}{c}{\cellcolor[HTML]{EFEFEF}\textit{Standard Pipeline}}\\
            \midrule
			\multirow{1}*{\textbf{1 $\rightarrow$ 2 $\rightarrow$ 3}} 
			& \textbf{62.78} & \textbf{47.58} & \textbf{56.03} \\
            \midrule
            \multicolumn{4}{c}{\cellcolor[HTML]{EFEFEF}\textit{Necessity of Progressive Training}}\\
            \midrule
			\multirow{1}*{1 $\&$ 2 $\&$ 3} 
		  & 60.04 ($\downarrow$ 4.36\%) & 44.95 ($\downarrow$ 5.53\%) & 53.33 ($\downarrow$ 4.82\%) \\
            \midrule
            \multicolumn{4}{c}{\cellcolor[HTML]{EFEFEF}\textit{The Role of Each Stage}}\\
            \midrule
			\multirow{1}*{2 $\rightarrow$ 3} 
			& 57.86 ($\downarrow$ 7.84\%) & 44.21 ($\downarrow$ 7.08\%) & 51.79 ($\downarrow$ 7.57\%) \\
			\multirow{1}*{1 $\rightarrow$ 3} 
			& 59.28 ($\downarrow$ 5.58\%) & 44.28 ($\downarrow$ 6.94\%) & 52.61 ($\downarrow$ 6.10\%) \\
			\multirow{1}*{1 $\rightarrow$ 2} 
			& 61.21 ($\downarrow$ 2.50\%) & 45.76 ($\downarrow$ 3.83\%) & 54.34 ($\downarrow$ 3.02\%) \\
            \midrule
            \multicolumn{4}{c}{\cellcolor[HTML]{EFEFEF}\textit{Sequential Order of Stages}}\\
            \midrule
			\multirow{1}*{2 $\rightarrow$ 1 $\rightarrow$ 3} 
			& \underline{61.29} ($\downarrow$ 2.37\%) & \underline{46.32} ($\downarrow$ 2.65\%) & \underline{54.64} ($\downarrow$ 2.48\%) \\
			\bottomrule
		\end{tabular}
	}
    \vspace{-0.2cm}
\end{table}

Ablation studies and in-depth analysis are performed on models to highlight the necessity of progressive multi-stage training and the contribution of each training stage, as well as to assess the impact of their sequential order and the influence of medical-specific reward design on Med-R$^3$, 
with results shown in \Cref{tab:ablation} and \Cref{tab:reward_ablation}.

\subsection{Necessity of Progressive Training}

We consolidated the reward functions from all stages throughout training to verify the importance of progressively optimizing the model’s retrieval and reasoning capabilities in a staged manner. 
The experimental results are presented in \Cref{tab:ablation} (\underline{1 $\&$ 2 $\&$ 3}), where we observe a performance drop of 4.82\% compared to our original multi-stage training strategy (\underline{1 $\rightarrow$ 2 $\rightarrow$ 3}). 
A key contributing factor to this decline 
is the inherent complexity and potential conflicts among diverse reward signals, 
i.e., the retrieval-focused and reasoning-oriented components impose different behavioral pressures on the model, leading to unstable policy updates in training. 
For instance, early in training, the model may not yet possess a sufficiently developed reasoning structure to effectively utilize or prioritize external knowledge retrieval, resulting in misaligned gradient signals and diminished learning efficiency.

\subsection{The Role of Each Stage}
\label{subsec:stage_ablation}

To evaluate the significance of each stage, we conduct three ablation experiments by individually removing Stage 1, 2, and 3. We then assess the performance on benchmarks, 
with results summarized in \Cref{tab:ablation}. 
To control for the influence of training data volume on model performance, we set the total number of training epochs at 3 to align with the main experiment, and allocate 1.5 epochs to each of the remaining two stages for training.

\begin{table}[tb] 
    \centering
    \caption{\label{tab:reward_ablation}
    Analysis of the reward design using \textbf{\textit{Qwen2.5-7B}}. 
    ``\textbf{w/o}'': \textit{removing} the reward component from the original metric. 
    We compute the \textit{average} performance of the evaluated models across in-domain and out-of-domain benchmarks. 
    The best and second best scores are in \textbf{bold} and \underline{underlined}.
    }
	\resizebox{1\linewidth}{!}{
		\begin{tabular}{c|c|c|c}
			\toprule
			Method & In-Domain & Out-of-Domain & Avg. \\ 
            \midrule
            \textbf{\ours} & \textbf{62.78} & \textbf{47.58} & \textbf{56.03} \\
            \midrule
			\multicolumn{4}{c}{\cellcolor[HTML]{EFEFEF}\textit{Analysis of $\,\mathcal{R}_{reasoning} = R_{semantic} + R_{statistic} + R_{logical}$}}\\
            \midrule
            w/o $R_{semantic}$ & \underline{61.78} ($\downarrow$1.59\%) & 46.04 ($\downarrow$3.24\%) & \underline{54.79} ($\downarrow$2.21\%) \\
			w/o $R_{statistic}$ & 59.00 ($\downarrow$6.02\%) & 44.55 ($\downarrow$6.37\%) & 52.58 ($\downarrow$6.16\%) \\
            w/o $R_{logical}$ & 60.95 ($\downarrow$2.91\%) & \underline{46.05} ($\downarrow$3.22\%) & 54.33 ($\downarrow$3.03\%) \\
            \midrule
			\multicolumn{4}{c}{\cellcolor[HTML]{EFEFEF}\textit{Analysis of $\,\mathcal{R}_{retrieval} = R_{quality} + R_{breadth}$}}\\
            \midrule
			w/o $R_{quality}$ & 61.37 ($\downarrow$2.25\%) & 45.23 ($\downarrow$4.94\%) & 54.20 ($\downarrow$3.27\%) \\
			w/o $R_{breadth}$ & 60.68 ($\downarrow$3.35\%) & 45.05 ($\downarrow$5.32\%) & 53.73 ($\downarrow$4.10\%) \\
			\bottomrule
		\end{tabular}
	}
    \vspace{-0.2cm}
\end{table}

\textbf{Removing Stage 1:} 
Stage 1 aims to enhance the reasoning capabilities of the models. 
As depicted in \Cref{tab:ablation} (\underline{2 $\rightarrow$ 3}), the absence of Stage 1 leads to a notable degradation of 7.57\% in overall model performance.
This reduction occurs because Stage 1 serves as the foundation for Stage 2. 
When the model lacks strong medical reasoning capabilities, it is unlikely to recognize when external knowledge is needed or to construct effective retrieval queries for obtaining supportive information. Consequently, the potential benefits offered by subsequent training stages are substantially compromised.

\textbf{Removing Stage 2:} 
Stage 2 builds upon the groundwork established in the previous Stage 1 to refine the generation of semantically precise and retrieval-efficient query terms, which aims to further augment the quality and utility of retrieved documents from the external knowledge base. 
As seen in \Cref{tab:ablation} (\underline{1 $\rightarrow$ 3}), after removing Stage 2, there is a slight decrease of 6.10\% in the accuracy compared to the corresponding model trained with all three stages.

\textbf{Removing Stage 3:} 
Stage 3 is designed to further enhance the synergy between the model’s retrieval and reasoning capabilities, thereby improving its overall end-to-end performance in solving medical problems. 
As observed in \Cref{tab:ablation} (\underline{1 $\rightarrow$ 2}), the performance of models declines by 3.02\% when Stage 3 is excluded. 
However, due to the presence of complete Stage 1 and 2, the gap remains close to that of the complete three-stage training model and relatively low.

\subsection{Sequential Order of Stages}

We swap Stage 1 and Stage 2 to assess their impact on models. As shown in \Cref{tab:ablation} (\underline{2 $\rightarrow$ 1 $\rightarrow$ 3}), exchanging Stage 1 and Stage 2 leads to a slight performance drop by 2.48\%.  
Our analysis of each stage demonstrates the robustness and rationale of our original training sequence, which first optimizes reasoning capabilities, followed by adaptive refinement of retrieval abilities. 
It stems from the fact that a solid reasoning foundation is essential to recognize retrieval needs during problem-solving. Only with this prerequisite can the model meaningfully improve its ability to generate retrieval queries that effectively capture relevant information.

\subsection{Analysis of Reward Design}
\label{subsec:reward_ablation}

We analyzed the components of the two reward metrics, 
$\mathcal{R}_{reasoning}$ and $\mathcal{R}_{retrieval}$, with results summarized in \Cref{tab:reward_ablation}. 
The findings indicate that the absence of $R_{statistic}$ in $\mathcal{R}_{reasoning}$ leads to a higher drop in model performance. 
This highlights that the comprehensive coverage of medical entities and relations plays a crucial role in the reasoning process, which is a key distinction from reasoning processes in mathematics and code. 
Furthermore, the removal of $R_{breadth}$ from $\mathcal{R}_{retrieval}$ also results in a notable decline in performance, underscoring the importance of evaluating the contribution of the retrieved documents to the reasoning process.

\section{Related Work}
\label{sec:related_work}

\textbf{LLMs for Medical Domain} \quad 
Large Language Models (LLMs) have been increasingly deployed in the medical field~\cite{zeng2020meddialog,gu2021domain,clusmann2023future}. 
Extensive studies have focused on the direct use of medical data for the pretraining or supervised fine-tuning of LLMs~\cite{singhal2023large,thirunavukarasu2023large}. 
Prominent open-source milestones include MEDITRON~\cite{chen2023meditron}, a scaling series of medical pretrained models with 7B and 70B parameters, and UltraMedical series~\cite{zhang2024ultramedical} fine-tuned on Llama-3 families~\cite{grattafiori2024llama}. 
However, solving medical problems requires structured multi-step reasoning~\cite{lucas2024reasoning,savage2024diagnostic}. 
Existing research indicates that the pretraining and SFT phases bias models towards memorizing established problem-solving pathways, diminishing their generalization capabilities when confronted with novel scenarios~\cite{havrilla2024teaching,lai2025med}. 
In contrast, 
empirical evidence suggests that 
the reinforcement learning (RL) phase is more conducive to cultivating a model’s cognitive abilities, particularly in domains that demand substantial logical reasoning, 
such as mathematics, coding, and medicine~\cite{jaech2024openai,guo2025deepseek,team2025kimi}. 
Nevertheless, unlike math and coding skills, the specific knowledge inherent in the medical domain is not always available within foundation models~\cite{singhal2023large,wang2023pre}. 
Therefore, establishing connections to the knowledge corpus to acquire external information is of paramount importance for medical scenarios.

\textbf{Reinforcement Learning in LLMs} \quad 
Compared to supervised fine-tuning (SFT), reinforcement learning (RL) provides an alternative by enabling emergent reasoning of models without explicit supervision~\cite{jaech2024openai,team2025kimi}. 
The GRPO~\cite{shao2024deepseekmath,guo2025deepseek} RL framework has proven to be highly effective in augmenting the reasoning abilities of LLMs through rule-based rewards. 
Efforts have been made to improve the medical reasoning capability in LLMs via the RL process, with notable works including HuatuoGPT-o1~\cite{chen2024huatuogpt}, 
Med-S$^3$~\cite{jiang2025meds}, and AlphaMed~\cite{alhamed2025}. 
However, they neglect the fact that specialized medical knowledge 
is not widely existent within the model's intrinsic knowledge. 
In such cases, the internal knowledge within models is insufficient, and 
the integration of external knowledge becomes crucial~\cite{wang2024jmlr}. 
While prior works focused on enhancing retrieval-augmented reasoning through RL in general-domain settings~\cite{chen2025learning,song2025r1}, the reward modeling strategies developed therein are not well adapted to the clinical scenario. 
To mitigate this, we systematically employ the RL process to jointly enhance models' retrieval and reasoning capabilities in the medical domain. 

\section{Discussion}
\label{sec:conclusion}

In this paper, we introduce Med-R$^3$, a novel progressive reinforcement learning framework aimed at enhancing the medical retrieval-augmented reasoning capabilities of models, and construct a medical dataset that contains reference reasoning trajectories for the RL process. 
Experimental results indicate that \ours achieves 
state-of-the-art training outcomes among open-sourced base and instruct models in medical scenarios. 
However, there are some limitations. 
The evaluation of model performance during training, e.g., the reward calculation, relies on frontier large language models (e.g., DeepSeek-V3), which may lead to the propagation of biases.





\bibliographystyle{ACM-Reference-Format}
\bibliography{MedR3_2025}

\clearpage

\appendix

\renewcommand{\contentsname}{Appendix}
\tableofcontents
\addtocontents{toc}{\protect\setcounter{tocdepth}{2}}


\section{Data Preprocessing Details}
\label{sec:appendix_data_preparation}

\subsection{Training Datasets Details}
\label{subsec:appendix_training_datasets}

We collected the following datasets for our training, covering both closed-set exam questions and rare disease diagnostic scenarios.  

\begin{itemize} 
    \item \textbf{MedQA-USMLE} and \textbf{MedQA-MCMLE}~\cite{jin2020disease}: The MedQA dataset is sourced from the professional medical board exams, including content in English, simplified Chinese, and traditional Chinese. In this study, we use the subsets from the US and Mainland China. MedQA-USMLE is based on 18 English medical textbooks, while MedQA-MCMLE is constructed from 33 simplified Chinese textbooks.
    \item \textbf{MedMCQA}~\cite{pal2022medmcqa}: The MedMCQA dataset is a large-scale collection of multiple-choice questions covering 2,400 health-related topics and 21 medical disciplines. The content spans medicine (such as endocrinology, infectious diseases, haematology, and respiratory medicine), surgery (including general surgery, endocrinological surgery, breast and vascular surgery), as well as radiology and biochemistry. The questions are derived from both real-world clinical scenarios and simulated examinations. 
    \item \textbf{RareArena}~\cite{THUMedInfo_RareArena}: This dataset is a comprehensive dataset curated for rare disease diagnosis, encompassing nearly 50,000 patient cases that cover more than 4,000 diseases. It features 2 task settings, \textit{Rare Disease Screening (RDS)} where patient records are truncated before any diagnostic tests are performed, and \textit{Rare Disease Confirmation (RDC)} where cases are truncated at the point of final diagnosis.
\end{itemize}

The illustration of the data construction pipeline can be found in \Cref{fig:data_construction}. 
After data filtering, we constructed a final training dataset consisting of 2,140 questions from MedQA-USMLE, 1,204 from MedQA-MCMLE, 6,748 from MedMCQA, and 429 from RareArena. The training process was structured into 3 stages, with each stage trained for one full epoch.

For the \textit{MedQA-USMLE} and \textit{MedQA-MCMLE} datasets, the original data is divided into three parts: train, dev, and test. We utilize the training part for RL, and merge the dev and test subsets for evaluation. 
Since the test split of \textit{MedMCQA} does not include ground truth labels, we use the training set for RL and reserve the development set for evaluation purposes. 
As for \textit{RareArena}, we manually partitioned the RareArena-RDC task set into training and test sets at an 8:2 ratio using random sampling. 
It is worth noting that due to the distribution of question complexity, the majority of data filtered out from MedQA and MedMCQA were classified as overly simplistic, whereas in the case of RareArena, most excluded samples were considered overly complex.

\begin{figure*}[t]
    \centering
    \includegraphics[width=1\textwidth]{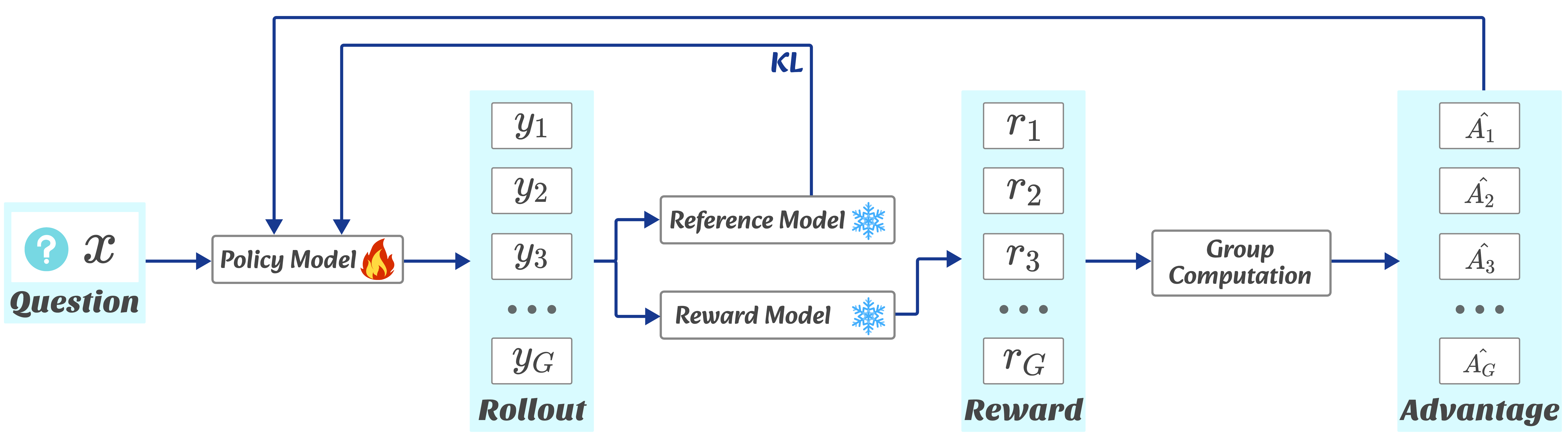}
    \caption{The Group Relative Policy Optimization (GRPO) pipeline.}
    \label{fig:GRPO}
\end{figure*}

\subsection{Knowledge Corpus Details}
\label{subsec:appendix_knowledge_corpus}

We have established a comprehensive medical knowledge base to support document retrieval during training and evaluation, which comprises four representative types of resources: \textit{academic papers}, \textit{entries}, \textit{books}, and \textit{guidelines}. 

\begin{itemize} 
    \item \textbf{Academic Papers} \quad Academic literature provide valuable insights drawn from recent scientific research, offering a strong theoretical basis for guiding clinical practice and public health policies. We sourced the publicly available academic articles from PubMed Central (PMC)~\cite{roberts2001pubmed}, and processed them following the pipeline of ~\citet{hakala-etal-2016-syntactic}.
    \item \textbf{Entries} \quad Medical entries serve as a rich source of multidimensional healthcare information, spanning clinical applications to biomedical research. We compile such entries by extracting and curating health-related content from the Wikipedia dataset~\cite{wikimedia_wikipedia_2024} to build the entries. The final collection comprises approximately 470k documents.
    \item \textbf{Books} \quad Medical textbooks are key sources of structured and validated medical knowledge, valuable for tackling complex clinical problems or staying updated on specific diseases. We compiled materials from the NCBI Bookshelf~\cite{sayers2021database}, and following the handling strategy of ~\citet{hakala-etal-2016-syntactic}, finally obtained 10k documents for knowledge retrieval. 
    \item \textbf{Guidelines} \quad Clinical practice guidelines play a critical role in Evidence-Based Medicine (EBM) by providing evide- nce-informed recommendations to guide clinical decision-making. We have incorporated the guideline data when \citet{chen2023meditron} training the MEDITRON series\footnote{https://huggingface.co/datasets/epfl-llm/guidelines}, and curated approximately 10k documents. 
\end{itemize}

The statistical information on volume and token number across knowledge resources can be found in \Cref{tab:medical_knowledge_corpus}. 
For retrieval, we segment the texts into chunks containing no more than 1,000 tokens. 
The segmentation prioritizes natural divisions such as chapters or sections. When such structural boundaries are unavailable or exceed the token limit, we apply truncation based on the predefined threshold to ensure consistency in input length. 

\section{Progressive Reinforcement Learning Details}

Here, we first introduce the core concepts and methodologies employed in our progressive reinforcement learning process, and then provide an elaboration on the algorithms and procedural significance across the different stages. 

\subsection{Preliminaries}

\subsubsection{Group Relative Policy Optimization (GRPO)}

During the training process, we utilize the Group Relative Policy Optimization (GRPO) as the RL algorithm. 
For each question $x \sim \mathcal{D}$, the behavior policy $\pi_{\theta_{\text{old}}}$ generates a set of $G$ candidate completions $\tau = \{y_i\}_{i=1}^{G} \sim \pi_{\theta_{\text{old}}}(\cdot|x)$,  
with each response receiving a scalar reward $r_i$. 
The training objective is to optimize the policy $\pi_{\theta}$ based on reference policy $\pi_{\theta_{\text{ref}}}$: 

\begin{equation}
\resizebox{\linewidth}{!}{
$\begin{split}
  \mathcal{J}(\theta) & = \mathbb{E}_{x \sim \mathcal{D}, \{y_i\}_{i=1}^{G} \sim \pi_{\theta_{\text{old}}}(\cdot|x)} \frac{1}{G} \sum_{i=1}^{G} [  
  \min ( \frac{\pi_{\theta}(y_i|x)}{\pi_{\theta_{\text{old}}}(y_i|x)} \hat{A_{i}}, \\ 
  & \text{clip} ( \frac{\pi_{\theta}(y_i|x)}{\pi_{\theta_{\text{old}}}(y_i|x)}, 1-\epsilon, 1+\epsilon ) \hat{A_{i}} ) - \beta \mathbb{D}_{\text{KL}} ( \pi_{\theta} || \pi_{\theta_{\text{ref}}} ) ]
\end{split}$
}
\label{eq:grpo}
\end{equation}
where the group-normalized advantage $\hat{A_{i}}$ of the $i$-th rollout in current group is defined as: 
\begin{equation*}
    \hat{A_{i}} = \frac{r_i - \text{mean}(\{r_j\}_{j=1}^{G})}{\text{std}(\{r_j\}_{j=1}^{G})}
    \label{eq:group_normalized_advantage}
\end{equation*}
Illustration of GRPO is shown in \Cref{fig:GRPO}. 
Here $\epsilon$ is the clipping ratio, a hyperparameter controlling the tolerance for policy deviation, and the \texttt{clip} function clips the importance weight $r_i$ to the interval $\left[1-\epsilon, 1+\epsilon \right]$, which stabilizes training and mitigating the risk of policy collapse. 
$\beta$ is the Kullback–Leibler (KL) loss coefficient~\cite{hall1987kullback}. 
To ensure stability during policy updates, a KL divergence penalty is included in the optimization objective, preventing significant deviations from the original reference policy models.

\subsubsection{Jaccard Similarity}

Jaccard similarity~\cite{jaccard1912distribution} is a widely used measure of similarity between two sets, quantifying the degree of overlap by comparing shared and distinct elements. 
It is defined as the ratio of the size of the intersection to the size of the union of the two sets. 
Formally, given two sets $A$ and $B$, the Jaccard similarity is expressed as: 
\begin{equation}
    Jaccard(A,B) =  \frac{|A \cap B|}{|A \cup B|}
    \label{eq:jaccard_similarity}
\end{equation}
where $|A \cap B|$ denotes the number of elements common to both sets, and $|A \cup B|$ represents the total number of distinct elements in either set. The Jaccard similarity ranges from 0, indicating no overlap, to 1, indicating identical sets. It is applicable to sets containing either numerical values or categorical strings. 
Here we use Jaccard similarity to compute the reward signals during Stage 1 of deliberative reasoner cultivation, specifically for quantifying the overlap in reasoning paths used to derive $R_{statistic}$ and $R_{logical}$.

\subsubsection{Evidence-Based Medicine (EBM)} 
\label{subsubsec:EBM_appendix}
It involves the integration of individual expertise with the most reliable external clinical evidence derived from systematic research~\cite{sackett1996evidence}. 
To identify the ``best evidence'', researchers evaluate trial quality using grading systems that assess the likelihood of bias and the reliability of results. The \textbf{\textit{hierarchy of evidence}} informs clinical decision-making by prioritizing evidence according to its methodological strength, as illustrated in \Cref{fig:hierarchy_of_evidence}. 
Its structure is typically organized from the highest to the lowest level of reliability, informing the quality scoring reward ($R_{quality}$) for the retrieved documents in \Cref{eq:authority}. 

\begin{figure*}[t]
    \centering
    \includegraphics[width=1\textwidth]{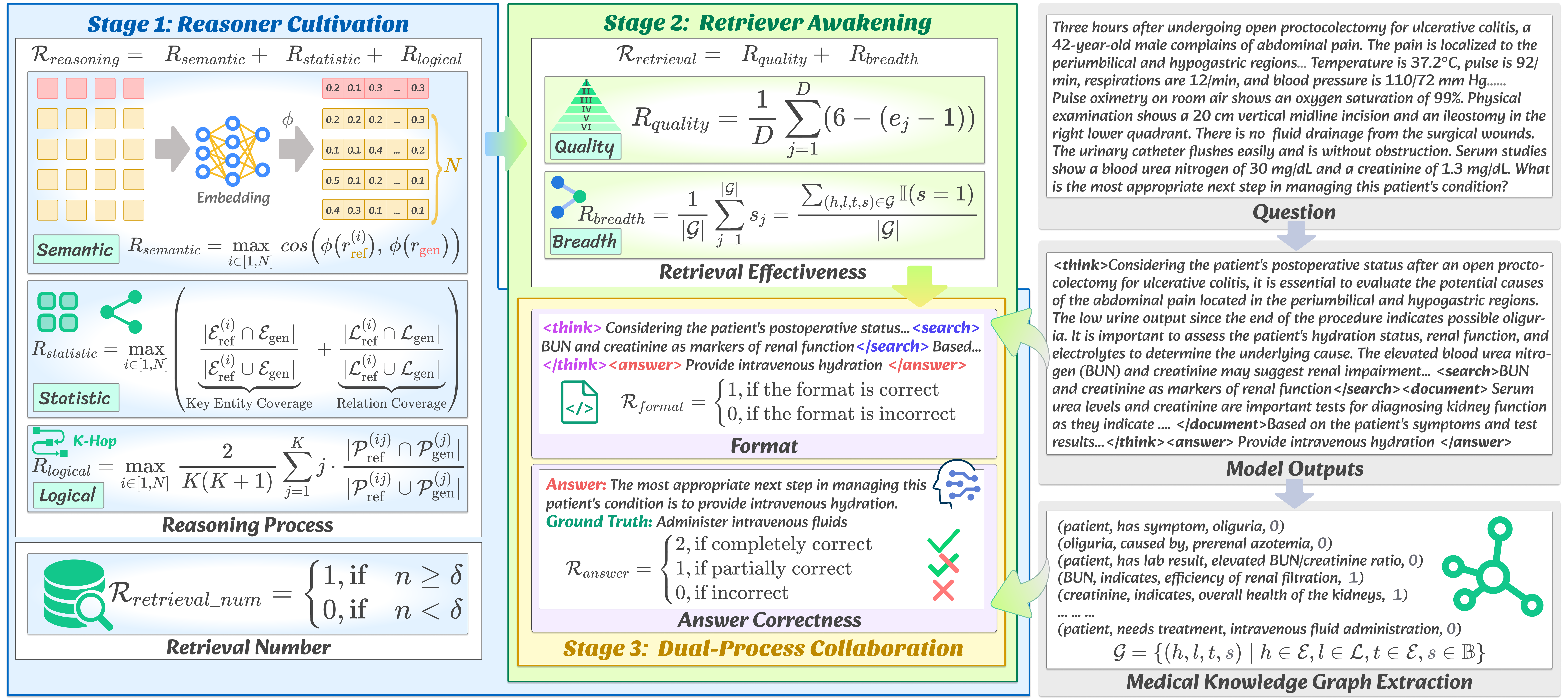}
    \caption{Detailed illustration of the reward function design during the progressive reinforcement learning of Med-R$^3$. 
    The model's natural language outputs are first structured into knowledge graph representations. In each training stage, tailored reward functions assess both the textual answer and the corresponding reasoning trajectory in the format of \Cref{eq:kg_retrieval_version}.}
    \label{fig:reinforcement_learning_appendix}
\end{figure*}                            

\begin{itemize} 
    \item \textbf{Systematic Reviews/Meta-Analyses (SR/MA)} represent \textit{the highest} level of evidence, evaluating the consistency and risk of bias across medical studies to summarize the overall effect of interventions or exposures. 
    \item \textbf{Randomized Controlled Trials (RCTs)} constitute \textit{the second-highest} level of evidence, designed to minimize confounding biases and assess causal relationships between interventions and outcomes across study groups. 
    \item \textbf{Cohort Studies} represent \textit{the third-highest} level of evidence. While both retrospective and prospective designs are subject to bias, prospective studies are generally more reliable, offering better control over information bias. 
    \item \textbf{Case-Control Studies} rank as \textit{the fourth-highest} level of evidence. They aim to identify associations between outcomes and past exposures but are prone to selection, information, and confounding biases, which limit their reliability compared to cohort studies. 
    \item \textbf{Individual Case Series and Case Reports} are of \textit{the second-lowest} level, essentially consisting of uncontrolled studies without a comparison group. The lack of controls limits the ability to establish reliable associations between interventions, exposures, or risk factors and outcomes.  
    \item \textbf{Expert Opinion} is considered \textit{the lowest} level of evidence due to its inherent susceptibility to bias. Experts may favor information that supports existing beliefs, leading to confirmation bias, conflicts of interest, and a narrow focus that overlooks broader contextual factors.
\end{itemize}

\subsection{Motivation and Principle of Reward Modeling}
\label{subsec:appendix_reward_modeling}

We provide an in-depth description of the reward computation process associated with progressive RL described in the main content. 
As shown in the right part of \Cref{fig:reinforcement_learning_appendix}, during the verification phase of each training stage, we first parse the model's natural language output into a structured knowledge graph representation in the form of \Cref{eq:kg_retrieval_version}. 
The stage-specific reward metric is then applied to assess both the \textit{natural language response} and the structural and logical accuracy of the \textit{reasoning trajectory encoded in the knowledge graph}.

\subsubsection{Stage 1: Reasoner Cultivation}

The primary objective of this stage is twofold: \textbf{\textit{\underline{(1)}}} to regularize the model's output format, ensuring structured and consistent responses, and \textbf{\textit{\underline{(2)}}} to cultivate its ability to perform logical reasoning when answering medical questions. In addition, by incorporating reward signals based on the number of knowledge retrievals, we encourage the model to make increased use of external medical knowledge sources, which lays the groundwork for enhancing the model's retrieval capabilities in the subsequent stage. 
The reward is composed of four aspects: \textit{format}, \textit{answer correctness}, \textit{reasoning process} and \textit{retrieval number}, as depicted in \Cref{fig:reinforcement_learning_appendix}. 
While format and answer correctness are commonly used reward components, we place particular emphasis on the rationale behind the design of rewards for reasoning process and retrieval number. 

\textbf{\textit{Reasoning Process.}} \quad 
In designing the reward function, we have considered the following aspects: 
\begin{itemize}[leftmargin=*]
    \item \textbf{Alignment with High-Quality Reasoning Trajectories}: We aim to guide the model's reasoning process toward high-quality trajectories observed in frontier models. To this end, our evaluation criteria incorporate semantic and structural characteristics of reference reasoning paths (see \Cref{eq:stage1_reasoning} to \Cref{eq:logical_score}), enabling the model to learn from expert-like inference patterns. 
    \item \textbf{Capturing Medical Reasoning Specifics}: Medical reasoning often requires both logical coherence and comprehensive coverage of relevant entities and their relationships. Unlike mathematical or programming tasks that emphasize rigid logical order, medical reasoning allows for multiple valid pathways, provided they are semantically rich and clinically grounded~\cite{wu2023medical,gao2025leveraging}. Accordingly, we introduce: \textit{\textbf{\underline{(1)}}} a structural coherence score based on the sequential logic of the reasoning path, as outlined in \Cref{eq:logical_score}, and \textit{\textbf{\underline{(2)}}} a completeness score evaluating the coverage of key entities and relations, as shown in \Cref{eq:statistic_score}. 
    \item \textbf{Robust Evaluation via Maximum Similarity Matching}: For each training instance, we provide multiple reference reasoning trajectories. However, these references may not be highly similar to one another due to the diversity of valid clinical reasoning. Therefore, in designing \Cref{eq:semantic_score} to \Cref{eq:logical_score}, we opt to use the maximum similarity score between the model's reasoning process and the reference reasoning processes, both in terms of semantics and structure, rather than an average score, to ensure robust and meaningful scoring.
    \item \textbf{Flexibility and Generalization}: Leveraging the strengths of reinforcement learning (RL), we avoid rigidly constraining the model's reasoning trajectory. Instead, we encourage it to approximate high-quality reasoning behaviors while allowing room for variation and creativity. This flexible guidance enhances the model's generalization capability when faced with novel or out-of-distribution reasoning tasks. To support this goal, we design a multi-dimensional reward scheme that captures various aspects of medical logic reasoning, as depicted in \Cref{eq:stage1_reasoning}. 
\end{itemize}

\begin{table}[t] 
    \centering
    \caption{\label{tab:retrieval_number_comparison}
    Comparison of average retrieval frequency with and without $\mathcal{R}_{retrieval\_num}$ in Stage 1.}
	\setlength{\tabcolsep}{1.5pt}
	\resizebox{1\linewidth}{!}{
		\begin{tabular}{c|c|ccc}
			\toprule
			\multirow{2}*{Model} & \multirow{2}*{Method} & \multicolumn{3}{c}{Average Retrieval Frequency} \\ 
            \cline{3-5}
            & & Stage 1 & Stage 2 & Stage 3 \\
			\midrule
			\multirow{2}*{Qwen2.5-7B} & w/o $\mathcal{R}_{retrieval\_num}$ & 0.44 & 0.83 & 0.92 \\
            & w/ $\mathcal{R}_{retrieval\_num}$ & 3.67 & 4.81 & 4.56 \\
			\bottomrule
		\end{tabular}
	}
\vspace{-0.5cm}
\end{table}

\textbf{\textit{Retrieval Number.}} \quad 
Inspired by the design in R1-Searcher~\cite{song2025r1}, 
one of the key motivations for incorporating this reward in Stage 1 is to lay the foundation for enhancing the model's retrieval capabilities in the subsequent stage. Specifically, only when the model develops the habit of querying external knowledge upon encountering unfamiliar concepts during reasoning, can there be meaningful room for optimizing its retrieval behavior. 
Here we have compared the average retrieval frequency with and without $\mathcal{R}_{retrieval\_num}$ in Stage 1, and conducted over-search detection.
\begin{itemize}
    \item \underline{\textit{Retrieval Frequency Comparison}}: 
    Without the reward signal $\mathcal{R}_{retrieval\_num}$ in Stage 1, certain models tend to rely solely on internal knowledge and cease performing external retrievals altogether. As observed in \Cref{tab:retrieval_number_comparison}, using Qwen2.5-7B as an example, the average number of knowledge retrievals during logical reasoning drops significantly when the retrieval count component $\mathcal{R}_{retrieval\_num}$ is omitted from the reward function.
    \item \underline{\textit{Over-Search Detection}}: We adopt the over-search detection method proposed in HiPRAG~\cite{wu2026hiprag}. 
    For each retrieval step in the model's reasoning trajectory, we extract the search query $q_i$ ($q_i$ is the content in the \texttt{<search>...</search>} tag pair), present it to the model $\mathcal{M}$ without the retrieved context (exclude the retrieved content in the \texttt{<document>...} \texttt{</document>} tag pair), and obtain a regenerated content $o'_i$. 
    We then use an external LLM judge (DeepSeek-V3) to assess the semantic equivalence between the original subsequent content 
    $o_i$ and the regenerated content $o'_i$. If they are semantically equivalent, it indicates that the retrieval step was redundant, 
    since the model could have reached the same conclusion without retrieval, and such step is flagged as \textit{over-search}. We employed the same detection prompt as HiPRAG~\cite{wu2026hiprag}, 
    and conducted evaluation under the following settings:
    \begin{itemize}
        \item Detected Model ($\mathcal{M}$): Qwen2.5-7B + Med-R$^3$
        \item LLM judge: DeepSeek-V3
    \end{itemize}
    These results indicate that the model does not suffer from severe over-searching. 
    Although the model retrieves 4.75 times per instance on average, only 4.45\% of retrieval steps are detected as redundant. 
    This suggests that most retrieval actions provide useful evidence rather than unnecessary context. 
    Moreover, harder benchmarks such as RareArena and MedXpertQA naturally require more retrieval steps, while their over-search rates remain within a narrow range. 
\end{itemize}

\begin{table}[t]
    \centering
    \caption{\label{tab:over-search-detection}
    Over-search detection across medical benchmarks. Here we employ Qwen2.5-7B + Med-R$^3$ as the detected model.}
    \setlength{\tabcolsep}{1.5pt}
    \resizebox{1\linewidth}{!}{
    \begin{tabular}{lccc}
        \toprule
        \textbf{Benchmark} & 
        \textbf{Avg. Retrieval Freq.} & 
        \textbf{Over-Search Rate(\%)} & 
        \textbf{Answer Acc.} \\
        \midrule
        MedQA-USMLE    & 3.28 & 3.84 & 68.64 \\
        MedQA-MCMLE    & 3.95 & 4.12 & 67.53 \\
        MedMCQA        & 4.26 & 4.20 & 68.97 \\
        RareArena-RDC  & 5.83 & 5.16 & 63.02 \\
        RareArena-RDS  & 5.12 & 5.40 & 45.76 \\
        MMLU-Med       & 4.10 & 3.54 & 75.81 \\
        NEJMQA         & 4.69 & 4.38 & 58.54 \\
        MedXpertQA     & 6.78 & 4.96 & 14.98 \\
        HealthBench    & 5.36 & 4.72 & 41.00 \\
        \midrule
        \textbf{Avg.}  & \textbf{4.82} & \textbf{4.48} & \textbf{56.03} \\
        \bottomrule
    \end{tabular}
    }
    \vspace{-0.5cm}
\end{table}

\subsubsection{Stage 2: Retriever Awakening}

The principal objective of this stage is to further enhance the model's ability to retrieve external knowledge built upon the medical reasoning capabilities developed in Stage 1. 
Specifically, we aim to improve the model's capacity to generate search queries such that the resulting documents  \textbf{\textit{\underline{(1)}}} exhibit higher semantic relevance, and \textbf{\textit{\underline{(2)}}} contribute more meaningfully to the logical reasoning process. 
In other words, this stage focuses on adaptively optimizing the content within the \texttt{<search>}...\texttt{</search>} tags.

The core reward metric in this phase is the \textit{retrieval effectiveness}, which evaluates both the quality of the retrieved documents and their influence within the model's overall reasoning trajectory, as outlined in \Cref{eq:stage2_reasoning}. 
For document quality assessment, we adopt the Evidence-Based Medicine (EBM) grading framework illustrated in \Cref{fig:hierarchy_of_evidence} to classify the retrieved documents into different quality levels. 
To evaluate the contribution of these retrieved documents to the model's reasoning trajectory, we compute the ratio of retrieved knowledge graph triples relative to the total number of triples in the reasoning trajectory, as defined in \Cref{eq:breadth}. This metric reflects the extent to which external knowledge contributes to the logical inference process.

\subsubsection{Stage 3: Dual-Process Collaboration}

After the previous two stages which have separately enhanced the model's reasoning and retrieval capabilities, 
in the final training stage, the model's end-to-end (E2E) performance on medical problem-solving is directly optimized based on the accuracy of the final answers. As a result, the reward function in this stage consists solely of \textit{format} and \textit{answer correctness}.


\section{Experiment Details}
\label{sec:appendix_experiment_details}

\subsection{Baselines}
\label{subsec:appendix_baseline}

Here we provide a comprehensive overview of the various models as well as methods that serve as baselines in our comparative analysis.

\begin{table*}[th!]
    \centering
    \caption{\label{tab:main_results_appendix} 
    Comprehensive comparison 1 of \ours with baselines, including close-sourced models, open-sourced medical-specific models, and open-sourced medical reasoning models, where we have provided the inference strategy for each competitor. 
    \colorbox[HTML]{DCE6FF}{Here we have selected the highest-performing one (marked with $^{\dagger}$) to represent the optimal} \colorbox[HTML]{DCE6FF}{performance of each method, as summarized in \Cref{tab:main_results}.} 
    $*$ denotes our re-implementation with the same amount of our constructed training data for a fair comparison. 
    $^\diamond$ represents the PRM guided Vote Sum (P-VS) strategy during inference, which unlocks the full potential of MedS$^3$-8B~\cite{jiang2025meds}. 
    The best and second best of each model are in \textbf{bold} and \underline{underlined}. 
} 
    \setlength{\tabcolsep}{0.6pt}
    \resizebox{\textwidth}{!}{
        \begin{tabular}{c|c|ccccccccccc}
            \toprule
            \multirow{2}{*}{Model} & \multirow{2}{*}{\makecell{Training\\Method}} & \multirow{2}{*}{\makecell{Inference\\Method}} & \multicolumn{1}{|c}{MedQA-US} & MedQA-MC & MedMCQA & RareArena-RDC & RareArena-RDS & \multicolumn{1}{|c}{MMLU-Med} & NEJMQA & MedXpertQA & \multicolumn{1}{c|}{HealthBench} & \multirow{2}{*}{\textbf{Avg.}} \\
            \cline{4-12}
             & & & \multicolumn{5}{|c}{\textit{In-Domain (ID)}} &  \multicolumn{4}{|c|}{\textit{Out-of-Domain (OOD)}} & \\
            \midrule
            \multicolumn{13}{c}{\cellcolor[HTML]{EFEFEF}\textbf{\textit{Close-Sourced Models}}} \\
            \midrule
            \multirow{4}{*}{GPT-4o-mini} & \multirow{4}{*}{--} & Naive & \multicolumn{1}{|c}{72.31} & 67.28 & 66.00 & 50.67 & 38.75 & \multicolumn{1}{|c}{74.60} & 55.08 & 17.84 & 23.80 & \multicolumn{1}{|c}{51.81} \\
             & & CoT & \multicolumn{1}{|c}{\textbf{75.61}} & \underline{68.32} & \underline{69.58} & 48.97 & \underline{42.35} & \multicolumn{1}{|c}{\underline{80.63}}  & \underline{57.40} & \underline{20.74} & \underline{26.40} & \multicolumn{1}{|c}{\underline{54.44}} \\
             & & RAG & \multicolumn{1}{|c}{73.30} & 68.18 & 66.82 & \underline{51.25} & 39.47 & \multicolumn{1}{|c}{77.11}  & 54.89 & 15.63 & 25.90 & \multicolumn{1}{|c}{52.51} \\
            & & CoT-RAG & \multicolumn{1}{|c}{\underline{74.45}} & \textbf{69.25} & \textbf{70.52} & \textbf{52.03} & \textbf{43.24} & \multicolumn{1}{|c}{\textbf{80.96}} & \textbf{57.87} & \textbf{21.58} & \textbf{28.50} & \multicolumn{1}{|c}{\textbf{55.38}} \\
            \midrule
            \multicolumn{13}{c}{\cellcolor[HTML]{EFEFEF}\textbf{\textit{Open-Sourced Medical-Specific Models}}} \\
            \midrule
            MEDITRON-7B & -- & CoT-RAG & \multicolumn{1}{|c}{48.67} & 44.28 & 46.78 & 37.97 & 21.60 & \multicolumn{1}{|c}{50.12} & 33.40 & 16.55 & 15.50 & \multicolumn{1}{|c}{34.98} \\
            UltraMedical3-8B & -- & CoT-RAG & \multicolumn{1}{|c}{61.57} & 52.42 & 61.82 & 40.76 & 28.54 & \multicolumn{1}{|c}{72.52} & 45.31 & 10.82 & 21.30 & \multicolumn{1}{|c}{43.90} \\
            UltraMedical3.1-8B & -- & CoT-RAG & \multicolumn{1}{|c}{66.86} & 58.45 & 65.73 & 43.83 & 32.39 & \multicolumn{1}{|c}{75.86} & 50.66 & 12.08 & 24.10 & \multicolumn{1}{|c}{47.77} \\
            MEDITRON-70B & -- & CoT-RAG & \multicolumn{1}{|c}{60.60} & 55.64 & 56.48 & 75.16 & 48.85 & \multicolumn{1}{|c}{70.53} & 65.33 & 18.72 & 31.20 & \multicolumn{1}{|c}{53.61} \\
            \midrule
            \multicolumn{13}{c}{\cellcolor[HTML]{EFEFEF}\textbf{\textit{Open-Sourced Medical Reasoning Models}}} \\
            \midrule
            \multirow{4}{*}{HuatuoGPT-o1-8B} & \multirow{4}{*}{--} & Naive & \multicolumn{1}{|c}{62.67} & 62.78 & 65.45 & 45.86 & 36.02 & \multicolumn{1}{|c}{70.80} & \underline{51.57} & 13.39 & 26.20 & \multicolumn{1}{|c}{48.30} \\
            & & CoT & \multicolumn{1}{|c}{\textbf{67.84}} & \underline{65.40} & \underline{70.88} & 46.42 & \underline{39.15} & \multicolumn{1}{|c}{\underline{74.36}} & 51.03 & \underline{14.17} & 29.30 & \multicolumn{1}{|c}{\underline{50.95}} \\
             & & RAG & \multicolumn{1}{|c}{64.52} & 63.75 & 66.76 & \underline{48.50} & 38.94 & \multicolumn{1}{|c}{72.18} & 50.63 & 13.97 & \textbf{32.10} & \multicolumn{1}{|c}{50.15} \\
             & & CoT-RAG$^{\dagger}$ & \multicolumn{1}{|c}{\underline{66.97}} & \textbf{66.15} & \textbf{72.45} & \textbf{49.76} & \textbf{41.59} & \multicolumn{1}{|c}{\textbf{74.52}} & \textbf{51.60} & \textbf{14.34} & \underline{31.80} & \multicolumn{1}{|c}{\textbf{52.13}} \\
            \midrule
            \multirow{5}{*}{MedS$^3$-8B} & \multirow{5}{*}{--} & Naive & \multicolumn{1}{|c}{58.14} & 60.24 & 60.20 & 44.60 & 33.79 & \multicolumn{1}{|c}{72.12} & \underline{53.61} & 12.01 & 25.80 & \multicolumn{1}{|c}{46.72} \\
            & & CoT & \multicolumn{1}{|c}{69.11} & 65.23 & 61.35 & 43.79 & 34.08 & \multicolumn{1}{|c}{74.36} & 52.17 & \textbf{13.64} & 29.20 & \multicolumn{1}{|c}{49.21} \\
             & & RAG & \multicolumn{1}{|c}{63.67} & 62.68 & 62.51 & \textbf{47.83} & \underline{37.14} & \multicolumn{1}{|c}{70.50} & 52.97 & 12.23 & 31.00 & \multicolumn{1}{|c}{48.95} \\
             & & CoT-RAG & \multicolumn{1}{|c}{\underline{71.59}} & \underline{66.84} & \underline{64.95} & 45.17 & 34.63 & \multicolumn{1}{|c}{\underline{75.38}} & 52.89 & 11.76 & \underline{32.40} & \multicolumn{1}{|c}{\underline{50.62}} \\
             & & P-VS$^{\diamond \dagger}$ & \multicolumn{1}{|c}{\textbf{73.51}} & \textbf{69.63} & \textbf{65.47} & \underline{46.85} & \textbf{37.72} & \multicolumn{1}{|c}{\textbf{78.75}} & \textbf{55.09} & \underline{12.50} & \textbf{33.20} & \multicolumn{1}{|c}{\textbf{52.52}} \\
            \midrule
            \multirow{4}{*}{AlphaMed-8B} & \multirow{4}{*}{--} & Naive & \multicolumn{1}{|c}{59.30} & 62.50 & 62.85 & 40.88 & 35.36 & \multicolumn{1}{|c}{70.63} & 50.72 & \underline{23.67} & 28.10 & \multicolumn{1}{|c}{48.22} \\
             & & CoT & \multicolumn{1}{|c}{\underline{63.82}} & \underline{64.49} & \underline{65.85} & 41.57 & 37.38 & \multicolumn{1}{|c}{\textbf{72.26}} & 50.93 & \textbf{25.43} & 31.40 & \multicolumn{1}{|c}{\underline{50.35}} \\
             & & RAG & \multicolumn{1}{|c}{62.70} & 64.17 & 63.96 & \underline{41.83} & \textbf{38.22} & \multicolumn{1}{|c}{\underline{71.85}} & \underline{51.04} & 20.75 & \textbf{32.80} & \multicolumn{1}{|c}{49.70} \\
             & & CoT-RAG$^{\dagger}$ & \multicolumn{1}{|c}{\textbf{64.06}} & \textbf{64.98} & \textbf{66.43} & \textbf{43.55} & \underline{38.14} & \multicolumn{1}{|c}{71.44} & \textbf{51.48} & 22.01 & \underline{32.60} & \multicolumn{1}{|c}{\textbf{50.52}} \\
             \midrule
             \multirow{4}{*}{MedResearcher-R1-32B} & \multirow{4}{*}{--} & Naive & \multicolumn{1}{|c}{72.86} & 74.90 & 70.65 & 57.24 & 47.10 & \multicolumn{1}{|c}{79.86} & 55.34 & 14.76 & 30.80 & \multicolumn{1}{|c}{55.95} \\
            & & CoT & \multicolumn{1}{|c}{\underline{75.30}} & \underline{77.42} & \underline{72.64} & 59.38 & 49.21 
            & \multicolumn{1}{|c}{\underline{82.30}} & 56.92 & 15.38 & 33.40 & \multicolumn{1}{|c}{57.99} \\
            & & RAG & \multicolumn{1}{|c}{74.88} & 76.95 & 72.10 & \underline{61.76} & \underline{50.41} 
            & \multicolumn{1}{|c}{81.65} & \underline{58.12} & \underline{15.63} & \textbf{36.20} 
            & \multicolumn{1}{|c}{\underline{58.63}} \\
            & & CoT-RAG$^{\dagger}$ & \multicolumn{1}{|c}{\textbf{76.42}} & \textbf{78.15} & \textbf{73.58} & \textbf{62.21} & \textbf{50.93} 
            & \multicolumn{1}{|c}{\textbf{83.14}} & \textbf{58.47} & \textbf{15.82} & \underline{35.70} 
            & \multicolumn{1}{|c}{\textbf{59.38}} \\
            \bottomrule
        \end{tabular}
    }
\end{table*}

\begin{itemize} 
    \item \textbf{Close-Sourced Models}: Close-sourced models are regarded as embodying the current peak performance across various capabilities of LLMs, and serve as the strongest baselines. Here we have selected GPT-4o-mini~\cite{hurst2024gpt} for comparison.
    \item \textbf{Open-Sourced Medical-Specific Models}: These models refer to domain-specific models that were trained specifically on medical data. We have selected MEDITRON-7B, MEDITRON-70B~\cite{chen2023meditron}, UltraMedical3-8B, and UltraMedical3.1-8B~\cite{zhang2024ultramedical} to represent the open-sourced medical-specific models as competitors to assess Med-R$^3$’s relative advantage. Specifically, the backbone model of UltraMedical3.1-8B is LLaMA3.1-8B-Instruct, which facilitates a more intuitive comparison with our method. 
    \item \textbf{Open-Sourced Medical Reasoning Models}: Inspired by the breakthrough of OpenAI o1, these models are capable of complex medical reasoning, applying o1-like methods to the medical field. We have selected HuatuoGPT-o1-8B~\cite{chen2024huatuogpt}, MedS$^3$-8B~\cite{jiang2025meds}, AlphaMed-8B~\cite{alhamed2025} and MedResearcher-R1-32B~\cite{yu2025medresearcher} as competitors. 
    HuatuoGPT-o1-8B improved models capabilities in medical reasoning by performing a two-stage SFT + RL training on verifiable medical problems. 
    MedS$^3$-8B equips the model with a self-evolution paradigm and proposes PRM-guided Vote-Sum (P-VS) strategy during inference to enhance long-chain reasoning capabilities in the medical domain. 
    AlphaMed-8B is trained exclusively on rule-based RL. 
    The backbone models of these competitors are all LLaMA-3.1-8B-Instruct, allowing for \textit{a more direct comparison with LLaMA-3.1-8B-Instruct + Med-R$^3$}. 
    MedResearcher-R1-32B is trained with knowledge-informed trajectory synthesis, SFT, and RL on the Qwen2.5-32B backbone model to perform expert-level medical question answering and multi-step research.
    \item \textbf{Naive Response}: It refers to the case where the model directly generates answers to the medical questions without training or retrieval from external knowledge bases. 
    \item \textbf{Supervised Fine-Tuning (SFT)}: The model undergoes SFT using our constructed training dataset, with reasoning trajectories produced by proprietary models (e.g., DeepSeek-V3), enriched with knowledge obtained through retrieval mechanisms. 
    During training, retrieved documents are with masked losses. 
    \item \textbf{General Retrieval-Augmented Reasoning RL}: We also compare our method with approaches that enhance the model's retrieval-augmented reasoning capabilities through RL in the general domain. 
    R1-Searcher~\cite{song2025r1} is a two-stage outcome-based RL method designed to enhance the model's capabilities of searching and integrating additional knowledge during the reasoning process. 
    To ensure a more equitable comparison, we replaced the training dataset and the knowledge corpus for retrieval with one that matches our experimental setup. In the first stage, we trained the model for 1 epoch, and in the second stage for 2 epochs, with a total number of training samples matching that of Med-R$^3$. 
    ReSearch~\cite{chen2025learning} considers search operations as integral components of the reasoning chain, and trains models to reason with search via RL without using any supervised data on reasoning steps. 
    Here we also train the model for 3 epochs to align with our experimental configuration. 
    The RL process is implemented by the Group Relative Policy Optimization (GRPO)~\cite{shao2024deepseekmath} algorithm. 
\end{itemize}

We evaluate these baselines under different inference methods, including \textbf{\textit{Naive}} (direct generation), \textbf{\textit{CoT}} (pure reasoning), \textbf{\textit{RAG}} (pure retrieval), and \textbf{\textit{CoT-RAG}} (interleaved reasoning and retrieval), as detailed in \Cref{tab:main_results_appendix} and \Cref{tab:main_results_appendix2}.

\begin{table*}[th!]
    \centering
    \caption{\label{tab:main_results_appendix2} 
    Comprehensive comparison 2 of \ours with baselines including open-sourced base models and open-sourced instruct models, where we have provided the inference strategy for each competitor. 
    \colorbox[HTML]{DCE6FF}{Here we have selected the highest-performing one (marked with $^{\dagger}$) to represent the optimal performance of each method, as} \colorbox[HTML]{DCE6FF}{summarized in \Cref{tab:main_results}.} 
    $*$ denotes our re-implementation with the same amount of our constructed training data for a fair comparison. 
    The best and second best of each model are in \textbf{bold} and \underline{underlined}. 
} 
    \setlength{\tabcolsep}{0.6pt}
    \resizebox{\textwidth}{!}{
        \begin{tabular}{c|c|ccccccccccc}
            \toprule
            \multirow{2}{*}{Model} & \multirow{2}{*}{\makecell{Training\\Method}} & \multirow{2}{*}{\makecell{Inference\\Method}} & \multicolumn{1}{|c}{MedQA-US} & MedQA-MC & MedMCQA & RareArena-RDC & RareArena-RDS & \multicolumn{1}{|c}{MMLU-Med} & NEJMQA & MedXpertQA & \multicolumn{1}{c|}{HealthBench} & \multirow{2}{*}{\textbf{Avg.}} \\
            \cline{4-12}
             & & & \multicolumn{5}{|c}{\textit{In-Domain (ID)}} &  \multicolumn{4}{|c|}{\textit{Out-of-Domain (OOD)}} & \\
            \midrule
            \multicolumn{13}{c}{\cellcolor[HTML]{EFEFEF}\textbf{\textit{Open-Sourced Base / Instruct Models}}} \\
            \midrule
            \multirow{11}{*}{\makecell{LLaMA3.1-8B\\-Instruct}} & \multirow{2}{*}{--} & Naive & \multicolumn{1}{|c}{31.16} & 41.45 & 30.02 & 41.85 & 22.16 & \multicolumn{1}{|c}{37.12} & 50.41 & 14.90 & 16.50 & \multicolumn{1}{|c}{31.73} \\
            & & CoT-RAG & \multicolumn{1}{|c}{33.80} & 41.69 & 35.75 & 43.12 & 25.99 & \multicolumn{1}{|c}{42.83} & 52.97 & 11.24 & 20.20 & \multicolumn{1}{|c}{34.18} \\
            \cmidrule{2-13}
             & SFT & CoT-RAG & \multicolumn{1}{|c}{61.39} & 62.10 & 63.27 & 44.83 & 35.20 & \multicolumn{1}{|c}{63.08} & 49.64 & 12.72 & 31.80 & \multicolumn{1}{|c}{47.11} \\
             \cmidrule{2-13}
             & R1-Searcher$*$ & CoT-RAG & \multicolumn{1}{|c}{60.28} & 60.92 & 63.54 & 42.87 & 38.65 & \multicolumn{1}{|c}{70.19}  & 53.81 & \underline{15.73} & 34.60 & \multicolumn{1}{|c}{48.95} \\
             \cmidrule{2-13}
             & ReSearch$*$ & CoT-RAG & \multicolumn{1}{|c}{62.76} & 66.03 & 66.25 & 46.35 & 38.44 & \multicolumn{1}{|c}{71.27} & 53.26 & 14.65 & 38.70 & \multicolumn{1}{|c}{50.86} \\
             \cmidrule{2-13}
             & \multirow{4}{*}{\textbf{\ours}} & Naive & \multicolumn{1}{|c}{67.17} & 62.35 & 71.06 & 52.06 & 41.20 & \multicolumn{1}{|c}{74.15} & 56.00 & 12.29 & 36.20 & \multicolumn{1}{|c}{52.50} \\
             & & CoT & \multicolumn{1}{|c}{69.43} & 65.58 & 73.78 & 52.63 & 43.87 & \multicolumn{1}{|c}{\underline{76.84}}  & 56.39 & 14.76 & 39.40 & \multicolumn{1}{|c}{54.74} \\
             & & RAG & \multicolumn{1}{|c}{\underline{70.09}} & \underline{67.56} & \underline{74.80} & \underline{55.00} & \underline{45.52} & \multicolumn{1}{|c}{76.25}  & \underline{58.10} & 13.94 & \underline{40.60} & \multicolumn{1}{|c}{\underline{55.76}} \\
             & & CoT-RAG$^{\dagger}$ & \multicolumn{1}{|c}{\textbf{75.91}} & \textbf{75.95} & \textbf{75.89} & \textbf{57.34} & \textbf{47.16} & \multicolumn{1}{|c}{\textbf{79.07}}  & \textbf{60.60} & \textbf{16.48} & \textbf{42.80} & \multicolumn{1}{|c}{\textbf{59.02}} \\
             \midrule
             \multirow{11}{*}{Qwen2.5-7B} & \multirow{2}{*}{--} & Naive & \multicolumn{1}{|c}{22.58} & 39.14 & 28.77 & 32.17 & 23.10 & \multicolumn{1}{|c}{44.45} & 41.48 & 11.59 & 14.60 & \multicolumn{1}{|c}{28.65} \\
             &  & CoT-RAG & \multicolumn{1}{|c}{27.45} & 37.33 & 28.21 & 30.80 & 24.85 & \multicolumn{1}{|c}{48.16} & 45.52 & 10.97 & 18.90 & 
             \multicolumn{1}{|c}{30.24} \\
             \cmidrule{2-13}
              & SFT & CoT-RAG & \multicolumn{1}{|c}{52.56} & 50.04 & 57.90 & 53.45 & 34.67 & \multicolumn{1}{|c}{56.94}  & 48.23 & 11.28 & 29.70 & \multicolumn{1}{|c}{43.86} \\
             \cmidrule{2-13}
              & R1-Searcher$*$ & CoT-RAG & \multicolumn{1}{|c}{56.78} & 49.70 & 58.35 & 53.69 & 33.27 & \multicolumn{1}{|c}{66.81} & 52.98 & 12.55 & 32.90 & \multicolumn{1}{|c}{46.34} \\
             \cmidrule{2-13}
             & ReSearch$*$ & CoT-RAG & \multicolumn{1}{|c}{62.47} & 60.24 & 63.11 & 55.95 & 34.68 & \multicolumn{1}{|c}{70.29} & 52.30 & 12.67 & 38.10 & \multicolumn{1}{|c}{49.98} \\
             \cmidrule{2-13}
             & \multirow{4}{*}{\textbf{\ours}} & Naive & \multicolumn{1}{|c}{60.82} & 55.60 & 62.28 & 59.98 & 37.56 & \multicolumn{1}{|c}{69.72} & 54.09 & 12.22 & 34.60 & \multicolumn{1}{|c}{49.65} \\
              & & CoT & \multicolumn{1}{|c}{64.48} & 62.28 & 63.74 & \underline{61.05} & 38.09 & \multicolumn{1}{|c}{\underline{73.73}} & 52.57 & \underline{13.85} & 37.80 & \multicolumn{1}{|c}{51.95} \\
             & & RAG & \multicolumn{1}{|c}{\underline{65.99}} & \underline{64.83} & \underline{66.18} & 60.86 & \underline{42.49} & \multicolumn{1}{|c}{71.04} & \underline{55.92} & 12.78 & \underline{39.40} & \multicolumn{1}{|c}{\underline{53.28}} \\
             & & CoT-RAG$^{\dagger}$ & \multicolumn{1}{|c}{\textbf{68.64}} & \textbf{67.53} & \textbf{68.97} & \textbf{63.02} & \textbf{45.76} & \multicolumn{1}{|c}{\textbf{75.81}} & \textbf{58.54} & \textbf{14.98} & \textbf{41.00} & \multicolumn{1}{|c}{\textbf{56.03}} \\
             \midrule
             \multirow{11}{*}{Qwen3-8B} 
            & \multirow{2}{*}{--} 
            & Naive 
            & \multicolumn{1}{|c}{38.43} & 45.62 & 35.25 & 36.54 & 21.81 
            & \multicolumn{1}{|c}{55.47} & 48.23 & 13.56 & 18.50 
            & \multicolumn{1}{|c}{34.82} \\
            & & CoT-RAG 
            & \multicolumn{1}{|c}{46.85} & 50.12 & 45.66 & 48.90 & 32.40 
            & \multicolumn{1}{|c}{62.10} & 52.05 & 13.20 & 25.40 
            & \multicolumn{1}{|c}{41.85} \\
            \cmidrule{2-13}
            & SFT 
            & CoT-RAG 
            & \multicolumn{1}{|c}{61.54} & 59.83 & 62.45 & 64.21 & 38.62 
            & \multicolumn{1}{|c}{66.54} & 50.36 & \underline{14.82} & 33.70 
            & \multicolumn{1}{|c}{50.23} \\
            \cmidrule{2-13}
            & R1-Searcher$*$ 
            & CoT-RAG 
            & \multicolumn{1}{|c}{59.82} & 61.24 & 64.15 & 62.53 & 40.42 
            & \multicolumn{1}{|c}{71.21} & 56.84 & 14.23 & 36.40 
            & \multicolumn{1}{|c}{51.87} \\
            \cmidrule{2-13}
            & ReSearch$*$ 
            & CoT-RAG 
            & \multicolumn{1}{|c}{64.63} & 62.54 & 65.32 & 63.81 & 41.63 
            & \multicolumn{1}{|c}{73.42} & 54.93 & 14.54 & 40.30 
            & \multicolumn{1}{|c}{53.46} \\
            \cmidrule{2-13}
            & \multirow{4}{*}{\textbf{\ours}} 
            & Naive 
            & \multicolumn{1}{|c}{70.45} & 69.10 & 72.20 & 70.36 & 48.10 
            & \multicolumn{1}{|c}{78.95} & 56.20 & 13.38 & 39.80 
            & \multicolumn{1}{|c}{57.62} \\
            & & CoT 
            & \multicolumn{1}{|c}{72.16} & 71.48 & 74.05 & 71.92 & 50.62 
            & \multicolumn{1}{|c}{\underline{80.76}} & 57.64 & 14.62 & 42.50 
            & \multicolumn{1}{|c}{59.53} \\
            & & RAG 
            & \multicolumn{1}{|c}{\underline{73.58}} & \underline{72.86} & \underline{74.88} & \underline{73.74} & \underline{51.80} 
            & \multicolumn{1}{|c}{80.12} & \underline{58.36} & 14.21 & \underline{43.30} 
            & \multicolumn{1}{|c}{\underline{60.32}} \\
            & & CoT-RAG$^{\dagger}$ 
            & \multicolumn{1}{|c}{\textbf{76.84}} & \textbf{74.92} & \textbf{76.53} & \textbf{75.62} & \textbf{53.24} 
            & \multicolumn{1}{|c}{\textbf{82.83}} & \textbf{59.42} & \textbf{15.14} & \textbf{44.80} 
            & \multicolumn{1}{|c}{\textbf{62.15}} \\
            \midrule
             \multirow{11}{*}{Qwen2.5-14B} & \multirow{2}{*}{--} & Naive & \multicolumn{1}{|c}{50.01} & 50.70 & 42.85 & 43.17 & 26.58 & \multicolumn{1}{|c}{71.60} & 45.63 & 11.06 & 22.40 & \multicolumn{1}{|c}{40.44} \\
             & & CoT-RAG & \multicolumn{1}{|c}{56.83} & 58.75 & 52.67 & 61.26 & 44.39 & \multicolumn{1}{|c}{73.48} & 46.02 & 11.14 & 30.80 & \multicolumn{1}{|c}{48.37} \\
             \cmidrule{2-13}
             & SFT & CoT-RAG & \multicolumn{1}{|c}{68.85} & 70.22 & 70.15 & 75.27 & 52.82 & \multicolumn{1}{|c}{75.81}  & 47.08 & 11.54 & 37.70 & \multicolumn{1}{|c}{56.61} \\
             \cmidrule{2-13}
             & R1-Searcher$*$ & CoT-RAG & \multicolumn{1}{|c}{69.20} & 71.75 & 68.45 & 76.32 & 54.05 & \multicolumn{1}{|c}{77.69} & 52.08 & 12.65 & 43.90 & \multicolumn{1}{|c}{58.45} \\
             \cmidrule{2-13}
             & ReSearch$*$ & CoT-RAG & \multicolumn{1}{|c}{69.52} & 74.05 & 72.20 & 75.67 & 53.54 & \multicolumn{1}{|c}{80.25} & 50.65 & 13.10 & 41.30 & \multicolumn{1}{|c}{58.92} \\
             \cmidrule{2-13}
             & \multirow{4}{*}{\textbf{\ours}} & Naive & \multicolumn{1}{|c}{72.98} & 73.40 & 71.78 & \textbf{78.80} & 55.45 & \multicolumn{1}{|c}{78.87} & 55.35 & 13.67 & 40.60 & \multicolumn{1}{|c}{60.10} \\
             & & CoT & \multicolumn{1}{|c}{72.30} & 74.82 & 70.44 & 76.95 & 56.37 & \multicolumn{1}{|c}{\underline{83.67}}  & \underline{59.81} & \underline{15.33} & 43.80 & \multicolumn{1}{|c}{61.50} \\
             & & RAG & \multicolumn{1}{|c}{\underline{76.58}} & \underline{77.75} & \underline{73.28} & \underline{78.36} & \underline{57.11} & \multicolumn{1}{|c}{81.49} & 56.64 & 14.01 & \underline{44.70} & \multicolumn{1}{|c}{\underline{62.21}} \\
             & & CoT-RAG$^{\dagger}$ & \multicolumn{1}{|c}{\textbf{78.01}} & \textbf{80.59} & \textbf{75.42} & 77.94 & \textbf{58.15} & \multicolumn{1}{|c}{\textbf{85.33}}  & \textbf{62.40} & \textbf{15.69} & \textbf{46.20} & \multicolumn{1}{|c}{\textbf{64.41}} \\
            \bottomrule
        \end{tabular}
    }
\end{table*}

\subsection{Benchmarks}

To evaluate the performance of our proposed \ours in both standard and real-world clinical scenarios of the medical domain, we have selected eight medical datasets, including the MedQA-USMLE, MedQA-MCMLE~\cite{jin2020disease}, MedMCQA~\cite{pal2022medmcqa}, MMLU-Med~\cite{hendryckstest2021}, RareArena\cite{THUMedInfo_RareArena}, MedXpertQA\cite{zuo2025medxpertqa}, HealthBench\cite{arora2025healthbench} and NEJMQA\cite{katz2024gpt}. 
Among these benchmarks, MedQA series, MedMCQA and MMLU-Med are \textit{standard question-answering tasks}, while RareArena, MedXpertQA, HealthBench and NEJMQA are related to \textit{real-world clinical scenarios}. 
We employ \textit{LLM-as-Judge} based on the proprietary model DeepSeek-V3~\cite{liu2024deepseek} to assess the correctness of the responses, then compute the accuracy scores as our evaluation metric. 

We have introduced within-domain tasks in the section of \textit{training datasets details}, where we use the training subsets for the RL process and test subsets (if no ground-truth labels provided, we use the development subsets) for \underline{in-domain} evaluation. 
In the following, we give descriptions of the remaining \underline{out-of-domain} benchmarks: 

\begin{itemize} 
    \item \textbf{MMLU-Med}~\cite{hendryckstest2021}: The MMLU-Med task is composed of six medical domains, anatomy, clinical knowledge, professional medicine, human genetics, college medicine, and college biology, which are extracted from the MMLU benchmark. 
    \item \textbf{NEJMQA}~\cite{katz2024gpt}: The NEJMQA dataset is constructed from clinical case challenges sourced from The Lancet and the New England Journal of Medicine, which is centered on diagnostic reasoning using patient symptom information. 
    \item \textbf{MedXpertQA}~\cite{zuo2025medxpertqa}: The MedXpertQA dataset incorporates specialty-specific evaluations and realistic clinical case questions derived from authentic medical practice. Here we utilize the \textit{text part} of this benchmark.
    \item \textbf{HealthBench}~\cite{arora2025healthbench}: The HealthBench dataset consists of 5,000 physician-designed multi-turn healthcare conversations evaluated against 48,562 doctor-written rubric criteria to measure LLM performance in realistic medical scenarios.
\end{itemize}

\subsection{More Experimental Observations}
\label{subsec:appendix:experimental_results}

The performance of each baseline reported in \Cref{tab:main_results} in the main content represents the best result achieved in four different inference strategies. The comprehensive scores across all inference strategies are provided in \Cref{tab:main_results_appendix} and \Cref{tab:main_results_appendix2}. 
Furthermore, 
we summarize the additional observations and insights derived from our experiments as follows: 

\textbf{Tailored reward design counts for medical scenarios.} 
Compared to general outcome-based approaches that enhance retrieval-augmented reasoning through reinforcement learning (RL), e.g., R1-Searcher and ReSearch, our \ours achieves an average improvement of 
17.51\% and 13.31\% 
in medical problem-solving. This performance gain is primarily attributed to our effective supervision of the model’s medical reasoning process and the design of reward functions specifically tailored to the medical domain.

\textbf{Retrieval and reasoning capabilities need to be cultivated.} When employing the \textit{CoT-RAG} inference method, the performance improvements across different models relative to the \textit{naive responses} vary significantly. 
Smaller-scale models such as Qwen2.5-7B and LLaMA3.1-8B-Instruct exhibit only modest gains under the \textit{CoT-RAG} setup, with average improvements of just 5.55\% and 7.72\%, respectively, in the absence of targeted training. In contrast, the larger Qwen2.5-14B model demonstrates a stronger adaptation to \textit{CoT-RAG}, achieving a more substantial improvement of 19.61\%. This indicates that lightweight models, stand to benefit greatly from dedicated training aimed at enhancing their retrieval-augmented reasoning capabilities in the medical domain.

\textbf{RL boosts generalizable medical performances.} 
We compare the performance of models after SFT and RL phase. Specifically, R1-Searcher, ReSearch, and \ours all employ RL-based training approaches. As shown in \Cref{tab:main_results_appendix2}, on the in-domain benchmarks, the performance gap between SFT and R1-Searcher/ReSearch is relatively small, with SFT underperforming by 0.59\% and 5.39\%, respectively. However, on out-of-domain tasks, particularly on MMLU-Med, the gap becomes much larger, with SFT lagging behind R1-Searcher and ReSearch by 9.21\% and 10.07\%, respectively. 
A primary reason for this discrepancy lies in the nature of SFT as a strongly supervised learning paradigm. Under this setting, the model tends to learn by directly imitating the given reasoning trajectories, effectively memorizing problem-solving patterns rather than truly learning the underlying reasoning process. Consequently, its generalization capability is notably limited when faced with novel or unseen scenarios. 
It is worth noting that in our SFT setup, the training data includes retrieval-augmented reasoning trajectories that incorporate externally retrieved knowledge, which mitigates the inherent limitations of SFT to some extent. Nevertheless, a clear performance gap still remains.

\subsection{Ablation Study and Analysis Details}

We report the original performance scores of the models on each benchmark during training stage ablation as well as reward design ablation, as depicted in \Cref{tab:ablation_detail} and \Cref{tab:reward_ablation_detail}, which are the detailed versions of \Cref{tab:ablation} and \Cref{tab:reward_ablation} in the main content.

\begin{table*}[ht] 
    \centering
    \caption{\label{tab:ablation_detail}
    \textbf{Original scores for each benchmark of the ablation study in \Cref{tab:ablation} on training stages and sequential order.} Order: sequential order of Stage 1, 2, and 3. Specifically, ``\underline{1 $\&$ 2 $\&$ 3}'' refers to a joint training configuration in which reward functions from all three stages are merged and optimized concurrently. 
    The best and second best scores of each model are in \textbf{bold} and \underline{underlined}. 
    We evaluate these models under the inference strategy of \textbf{\textit{CoT-RAG}} (interleaved reasoning and retrieval). 
    }
	\setlength{\tabcolsep}{1pt}
	\resizebox{1\linewidth}{!}{
		\begin{tabular}{c|ccccccccccc}
            \toprule
            \multirow{2}{*}{Order} & \multirow{2}{*}{Model} & \multicolumn{1}{|c}{MedQA-US} & MedQA-MC & MedMCQA & RareArena-RDC & RareArena-RDS & \multicolumn{1}{|c}{MMLU-Med} & NEJMQA & MedXpertQA & \multicolumn{1}{c|}{HealthBench} & \multirow{2}{*}{\textbf{Avg.}} \\
            \cline{3-11}
             & & \multicolumn{5}{|c}{\textit{In-Domain (ID)}} &  \multicolumn{4}{|c|}{\textit{Out-of-Domain (OOD)}} &  \\
            \midrule
			\multicolumn{12}{c}{\cellcolor[HTML]{EFEFEF}\textit{\textbf{Standard Pipeline}}}\\
			\midrule
			\multirow{3}*{\makecell*[c]{\shortstack{1 $\rightarrow$ 2 $\rightarrow$ 3}}} 
			& Qwen2.5-7B & \multicolumn{1}{|c}{\textbf{68.64}} & \textbf{67.53} & \textbf{68.97} & 63.02 & \textbf{45.76} & \multicolumn{1}{|c}{\textbf{75.81}} & \textbf{58.54} & \textbf{14.98} & \textbf{41.00} & \multicolumn{1}{|c}{\textbf{56.03}} \\
			& LLaMA3.1-8B-Instruct & \multicolumn{1}{|c}{\textbf{75.91}} & \textbf{75.95} & \textbf{75.89} & \textbf{57.34} & \textbf{47.16} & \multicolumn{1}{|c}{\textbf{79.07}} & \textbf{60.60} & \underline{16.48} & \textbf{42.80} & \multicolumn{1}{|c}{\textbf{59.02}} \\
            & Qwen2.5-14B & \multicolumn{1}{|c}{\textbf{78.01}} & \textbf{80.59} & \underline{75.42} & \textbf{77.94} & \textbf{58.15} & \multicolumn{1}{|c}{\textbf{85.33}} & \textbf{62.40} & \textbf{15.69} & \textbf{46.20} & \multicolumn{1}{|c}{\textbf{64.41}} \\
            \midrule
			\multicolumn{12}{c}{\cellcolor[HTML]{EFEFEF}\textit{\textbf{Necessity of Progressive Training}}}\\
			\midrule
			\multirow{3}*{1 $\&$ 2 $\&$ 3} 
			& Qwen2.5-7B & \multicolumn{1}{|c}{64.83} & 64.18 & 65.26 & 62.87 & 43.04 & \multicolumn{1}{|c}{72.80} & 54.35 & 13.85 & 38.80 & \multicolumn{1}{|c}{53.33} \\
			& LLaMA3.1-8B-Instruct & \multicolumn{1}{|c}{70.67} & 71.85 & 70.79 & \underline{55.12} & 44.78 & \multicolumn{1}{|c}{75.43} & 58.64 & 15.34 & 40.60 & \multicolumn{1}{|c}{55.91} \\
            & Qwen2.5-14B & \multicolumn{1}{|c}{75.30} & 79.04 & 73.38 & 77.00 & 55.89 & \multicolumn{1}{|c}{\underline{84.65}} & 60.16 & \underline{15.42} & 44.20 & \multicolumn{1}{|c}{62.78} \\
			\midrule
			\multicolumn{12}{c}{\cellcolor[HTML]{EFEFEF}\textit{\textbf{Effectiveness of Stage 1 (Reasoner Cultivation)}}}\\
			\midrule
			\multirow{3}*{2 $\rightarrow$ 3} 
			& Qwen2.5-7B & \multicolumn{1}{|c}{62.92} & 60.96 & 63.78 & 61.57 & 40.05 & \multicolumn{1}{|c}{71.12} & 54.87 & 14.14 & 36.70 & \multicolumn{1}{|c}{51.79} \\
			& LLaMA3.1-8B-Instruct & \multicolumn{1}{|c}{66.85} & 69.36 & 68.95 & 51.11 & 43.80 & \multicolumn{1}{|c}{73.62} & 55.84 & 14.85 & 37.80 & \multicolumn{1}{|c}{53.58} \\
            & Qwen2.5-14B & \multicolumn{1}{|c}{72.69} & 74.81 & 72.81 & 75.33 & 56.95 & \multicolumn{1}{|c}{80.42} & 57.04 & 14.33 & 41.70 & \multicolumn{1}{|c}{60.68} \\
			\midrule
			\multicolumn{12}{c}{\cellcolor[HTML]{EFEFEF}\textit{\textbf{Effectiveness of Stage 2 (Retriever Awakening)}}}\\
			\midrule
			\multirow{3}*{1 $\rightarrow$ 3} 
			& Qwen2.5-7B & \multicolumn{1}{|c}{65.71} & 63.52 & 66.15 & 60.14 & 40.88 & \multicolumn{1}{|c}{71.69} & 53.96 & 13.97 & 37.50 & \multicolumn{1}{|c}{52.61} \\
			& LLaMA3.1-8B-Instruct & \multicolumn{1}{|c}{\underline{72.58}} & 72.43 & 73.02 & 53.17 & 42.19 & \multicolumn{1}{|c}{75.80} & 57.28 & 13.62 & 39.20 & \multicolumn{1}{|c}{55.48} \\
            & Qwen2.5-14B & \multicolumn{1}{|c}{74.80} & 76.74 & 74.62 & \underline{77.68} & 56.07 & \multicolumn{1}{|c}{82.58} & 57.83 & 14.95 & 42.80 & \multicolumn{1}{|c}{62.01} \\
			\midrule
			\multicolumn{12}{c}{\cellcolor[HTML]{EFEFEF}\textit{\textbf{Effectiveness of Stage 3 (Dual-Process Collaboration)}}} \\
			\midrule
			\multirow{3}*{1 $\rightarrow$ 2} 
			& Qwen2.5-7B & \multicolumn{1}{|c}{\underline{66.84}} & 65.45 & 66.94 & \underline{63.18} & \underline{43.63} & \multicolumn{1}{|c}{73.65} & \underline{55.63} & 14.56 & 39.20 & \multicolumn{1}{|c}{54.34} \\
			& LLaMA3.1-8B-Instruct & \multicolumn{1}{|c}{69.73} & 70.67 & 71.56 & 54.28 & 45.14 & \multicolumn{1}{|c}{\underline{78.23}} & 59.32 & \textbf{16.80} & 40.80 & \multicolumn{1}{|c}{56.28} \\
            & Qwen2.5-14B & \multicolumn{1}{|c}{76.55} & \underline{80.28} & 74.03 & 76.97 & 57.46 & \multicolumn{1}{|c}{83.26} & 59.47 & 14.78 & 44.50 & \multicolumn{1}{|c}{63.03} \\
			\midrule
			\multicolumn{12}{c}{\cellcolor[HTML]{EFEFEF}\textit{\textbf{Sequential Order of Stages}}}\\
			\midrule
			\multirow{3}*{2 $\rightarrow$ 1 $\rightarrow$ 3} 
			& Qwen2.5-7B & \multicolumn{1}{|c}{65.98} & \underline{65.78} & \underline{67.52} & \textbf{64.46} & 42.72 & \multicolumn{1}{|c}{\underline{74.90}} & 55.28 & \underline{14.80} & \underline{40.30} & \multicolumn{1}{|c}{\underline{54.64}} \\
			& LLaMA3.1-8B-Instruct & \multicolumn{1}{|c}{71.49} & \underline{73.07} & \underline{73.60} & 53.26 & \underline{45.98} & \multicolumn{1}{|c}{77.44} & \underline{60.09} & 15.93 & \underline{42.10} & \multicolumn{1}{|c}{\underline{57.00}} \\
            & Qwen2.5-14B & \multicolumn{1}{|c}{\underline{77.45}} & 78.57 & \textbf{75.59} & 76.54 & \underline{57.72} & \multicolumn{1}{|c}{82.91} & \underline{60.75} & 15.16 & \underline{45.50} & \multicolumn{1}{|c}{\underline{63.35}} \\
			\bottomrule
		\end{tabular}
	}
\end{table*}

\begin{table*}[ht] 
    \centering
    \caption{\label{tab:reward_ablation_detail}
    \textbf{Original scores for each benchmark of the ablation study on reward design in \Cref{tab:reward_ablation}.} \textbf{w/o}: \textit{removing} the reward component from the original metric. The best and second best scores are marked in \textbf{bold} and \underline{underlined}. 
    We employ \textbf{\textit{Qwen2.5-7B}} for analysis, and evaluate the model under the inference strategy of \textbf{\textit{CoT-RAG}} (interleaved reasoning and retrieval). 
    }
	\setlength{\tabcolsep}{2pt}
	\resizebox{1\linewidth}{!}{
		\begin{tabular}{ccccccccccc}
            \toprule
            \multirow{2}{*}{Method} & \multicolumn{1}{|c}{MedQA-US} & MedQA-MC & MedMCQA & RareArena-RDC & RareArena-RDS & \multicolumn{1}{|c}{MMLU-Med} & NEJMQA & MedXpertQA & \multicolumn{1}{c|}{HealthBench} & \multirow{2}{*}{\textbf{Avg.}} \\
            \cline{2-10}
             & \multicolumn{5}{|c}{\textit{In-Domain (ID)}} &  \multicolumn{4}{|c|}{\textit{Out-of-Domain (OOD)}} & \\
            \midrule
            Naive Response 
            & \multicolumn{1}{|c}{22.58} & 39.14 & 28.77 & 32.17 & 23.10 
            & \multicolumn{1}{|c}{44.45} & 41.48 & 11.59 
            & 14.60 & \multicolumn{1}{|c}{28.65} \\
            \midrule
            \multicolumn{11}{c}{\cellcolor[HTML]{EFEFEF}\textit{\textbf{Analysis of $\,\mathcal{R}_{reasoning} = R_{semantic} + R_{statistic} + R_{logical}$}}}\\
            \midrule
            w/o $R_{semantic}$ 
            & \multicolumn{1}{|c}{66.47} & \underline{67.19} & 67.25 & \textbf{64.33} & 43.68 
            & \multicolumn{1}{|c}{\underline{75.04}} & 55.36 & \underline{14.86} 
            & \underline{38.90} & \multicolumn{1}{|c}{\underline{54.79}} \\
            w/o $R_{statistic}$ 
            & \multicolumn{1}{|c}{64.06} & 63.34 & 64.80 & 61.28 & 41.53 
            & \multicolumn{1}{|c}{72.47} & 55.91 & 13.72 
            & 36.10 & \multicolumn{1}{|c}{52.58} \\
            w/o $R_{logical}$ 
            & \multicolumn{1}{|c}{64.95} & 65.81 & 66.84 & 62.67 & \underline{44.46} 
            & \multicolumn{1}{|c}{74.28} & \underline{57.42} & 14.30 
            & 38.20 & \multicolumn{1}{|c}{54.33} \\
            \midrule
            \multicolumn{11}{c}{\cellcolor[HTML]{EFEFEF}\textit{\textbf{Analysis of $\,\mathcal{R}_{retrieval} = R_{quality} + R_{breadth}$}}}\\
            \midrule
            w/o $R_{quality}$ 
            & \multicolumn{1}{|c}{\underline{68.18}} & 66.07 & \underline{67.35} & 62.46 & 42.79 
            & \multicolumn{1}{|c}{73.82} & 54.73 & 14.67 
            & 37.70 & \multicolumn{1}{|c}{54.20} \\
            w/o $R_{breadth}$ 
            & \multicolumn{1}{|c}{66.35} & 64.84 & 66.92 & 61.70 & 43.58 
            & \multicolumn{1}{|c}{73.40} & 55.34 & 14.06 
            & 37.40 & \multicolumn{1}{|c}{53.73} \\
            \midrule
            \textbf{\ours} 
            & \multicolumn{1}{|c}{\textbf{68.64}} & \textbf{67.53} & \textbf{68.97} & \underline{63.02} & \textbf{45.76} 
            & \multicolumn{1}{|c}{\textbf{75.81}} & \textbf{58.54} & \textbf{14.98} 
            & \textbf{41.00} & \multicolumn{1}{|c}{\textbf{56.03}} \\
			\bottomrule
		\end{tabular}
	}
\end{table*}

\subsection{Environment and Hardware Configurations}

The experiment utilizes the following core libraries and their respective versions: torch=2.5.1, CUDA\_version=12.4, ray=2.40.0, vllm=0.7.3, verl=0.2.0.post2, transfomrers=4.49.0, datasets=3.3.2, tqdm=4.40.0, flash-attn=2.5.8, pyarrow=19.0.1, tensordict=0.5.0. 
Experiments are conducted using 32 NVIDIA 
GPUs with 96GB memory.

\begin{table*}[t]
    \centering
    \caption{\label{tab:cross_scorer_judge}
    Robustness analysis of Med-R$^3$ under open-weight trajectory scoring and cross-judge evaluation.
    Here we employ \textbf{\textit{Qwen2.5-7B}} as the backbone model.
    $*$ denotes our re-implementation with the same amount of our constructed training data for a fair comparison. 
    The trajectory scorer is used during training data construction, while the final evaluator is used only for LLM-as-Judge during test-time evaluation. 
    The best and second best scores of each model are in \textbf{bold} and \underline{underlined}. 
    }
    \setlength{\tabcolsep}{1.5pt}
    \resizebox{1\linewidth}{!}{
    \begin{tabular}{c|c|c|ccccc|cccc|c}
        \toprule
        \multirow{2}{*}{\makecell{Trajectory\\Scorer}} & \multirow{2}{*}{\makecell{Final\\Evaluator}} & \multirow{2}{*}{Method} & MedQA-US & MedQA-MC & MedMCQA & RareArena-RDC & RareArena-RDS & MMLU-Med & NEJMQA & MedXpertQA & HealthBench & \multirow{2}{*}{\textbf{Avg.}} \\
        \cline{4-12}
         & & & \multicolumn{5}{c}{\textit{In-Domain (ID)}} & \multicolumn{4}{|c|}{\textit{Out-of-Domain (OOD)}} &  \\
        \midrule
        \multirow{10}{*}{DeepSeek-V3} & \multirow{5}{*}{DeepSeek-V3}
        & Naive Response 
        & 22.58 & 39.14 & 28.77 & 32.17 & 23.10 & 44.45 & 41.48 & 11.59 & 14.60 & 28.65 \\
        & & SFT            
        & 52.56 & 50.04 & 57.90 & 53.45 & 34.67 & 56.94 & 48.23 & 11.28 & 29.70 & 43.86 \\
         & & R1-Searcher$^{*}$ 
        & 56.78 & 49.70 & 58.35 & 53.69 & 33.27 & 66.81 & \underline{52.98} & 12.55 & 32.90 & 46.34 \\
         & & ReSearch$^{*}$ 
        & \underline{62.47} & \underline{60.24} & \underline{63.11} & \underline{55.95} & \underline{34.68} & \underline{70.29} & 52.30 & \underline{12.67} & \underline{38.10} & \underline{49.98} \\
         & & \textbf{Med-R$^3$ (ours)}          
        & \textbf{68.64} & \textbf{67.53} & \textbf{68.97} & \textbf{63.02} & \textbf{45.76} 
        & \textbf{75.81} & \textbf{58.54} & \textbf{14.98} & \textbf{41.00} & \textbf{56.03} \\

        \cline{2-13}

         & \multirow{5}{*}{\makecell{LLaMA3.1-70B\\-Instruct}}
        & Naive Response 
        & 22.13 & 38.51 & 28.24 & 31.68 & 22.54 & 43.71 & 40.86 & 11.18 & 14.23 & 28.12 \\
        & & SFT            
        & 51.87 & 49.42 & 57.16 & 52.82 & 34.03 & 56.08 & 47.61 & 10.94 & 29.08 & 43.22 \\
         & & R1-Searcher$^{*}$ 
        & 55.92 & 49.06 & 57.64 & 53.14 & 32.74 & 65.98 & \underline{52.27} & 12.18 & 32.57 & 45.72 \\
         & & ReSearch$^{*}$    
        & \underline{61.76} & \underline{59.43} & \underline{62.36} & \underline{55.07} & \underline{34.26} & \underline{69.42} & 51.69 & \underline{12.33} & \underline{37.62} & \underline{49.33} \\
         & & \textbf{Med-R$^3$ (ours)}          
        & \textbf{67.72} & \textbf{66.58} & \textbf{68.01} & \textbf{62.13} & \textbf{44.73} 
        & \textbf{74.62} & \textbf{57.59} & \textbf{14.21} & \textbf{40.63} & \textbf{55.14} \\

        \midrule

        \multirow{10}{*}{\makecell{LLaMA3.1-70B\\-Instruct}} & \multirow{5}{*}{DeepSeek-V3}
        & Naive Response 
        & 22.58 & 39.14 & 28.77 & 32.17 & 23.10 & 44.45 & 41.48 & 11.59 & 14.60 & 28.65 \\
        & & SFT            
        & 52.12 & 49.58 & 57.36 & 52.91 & \underline{34.18} & 56.31 & 47.79 & 11.04 & 29.16 & 43.38 \\
         & & R1-Searcher$^{*}$ 
        & 56.24 & 49.18 & 57.81 & 53.08 & 32.85 & 66.23 & \underline{52.41} & 12.31 & 32.36 & 45.83 \\
         & & ReSearch$^{*}$    
        & \underline{61.93} & \underline{59.62} & \underline{62.52} & \underline{55.28} & 34.07 & \underline{69.61} & 51.76 & \underline{12.46} & \underline{37.48} & \underline{49.41} \\
         & & \textbf{Med-R$^3$ (ours)}          
        & \textbf{68.03} & \textbf{66.91} & \textbf{68.22} & \textbf{62.26} & \textbf{44.89} 
        & \textbf{75.17} & \textbf{57.86} & \textbf{14.54} & \textbf{40.37} & \textbf{55.36} \\

        \cline{2-13}

         & \multirow{5}{*}{\makecell{LLaMA3.1-70B\\-Instruct}}
         & Naive Response 
        & 22.13 & 38.51 & 28.24 & 31.68 & 22.54 & 43.71 & 40.86 & 11.18 & 14.23 & 28.12 \\
        & & SFT            
        & 51.32 & 48.91 & 56.61 & 52.19 & 33.47 & 55.47 & 47.05 & 10.78 & 28.51 & 42.70 \\
         & & R1-Searcher$^{*}$ 
        & 55.37 & 48.52 & 57.03 & 52.48 & 32.19 & 65.27 & \underline{51.68} & 11.98 & 31.89 & 45.16 \\
         & & ReSearch$^{*}$ 
        & \underline{61.15} & \underline{58.82} & \underline{61.77} & \underline{54.43} & \underline{33.61} & \underline{68.74} & 51.02 & \underline{12.11} & \underline{36.94} & \underline{48.73} \\
         & & \textbf{Med-R$^3$ (ours)}          
        & \textbf{66.96} & \textbf{65.84} & \textbf{67.28} & \textbf{61.29} & \textbf{43.92} 
        & \textbf{73.91} & \textbf{56.92} & \textbf{13.92} & \textbf{39.94} & \textbf{54.44} \\
        \bottomrule
    \end{tabular}
    }
\end{table*}

\begin{table*}[th] 
    \centering
    \caption{\label{tab:meta-analysis}
    \textbf{Meta-evaluation results across different evaluation strategies, including the exact match judge and the proprietary model \textit{DeepSeek-V3}'s judge.} We sample 30 instances per medical dataset.
    }
	\setlength{\tabcolsep}{1pt}
	\resizebox{1\linewidth}{!}{
		\begin{tabular}{c|ccccccccccc}
            \toprule
            \multirow{2}{*}{\makecell{Backbone\\Model}} & \multirow{2}{*}{\makecell{Evaluation\\Strategy}} & \multicolumn{1}{|c}{MedQA-US} & MedQA-MC & MedMCQA & RareArena-RDC & RareArena-RDS & \multicolumn{1}{|c}{MMLU-Med} & NEJMQA & MedXpertQA & \multicolumn{1}{c|}{HealthBench} & \multirow{2}{*}{\textbf{Avg.}} \\
            \cline{3-11}
             & & \multicolumn{5}{|c}{\textit{In-Domain (ID)}} & \multicolumn{4}{|c|}{\textit{Out-of-Domain (OOD)}} &  \\
            \midrule
            \multirow{2}*{Qwen2.5-7B} 
            & Exact Match 
            & \multicolumn{1}{|c}{0.80 (24/30)} & 0.70 (21/30) & 0.73 (22/30) & 0.30 (9/30) & 0.37 (11/30) 
            & \multicolumn{1}{|c}{0.60 (18/30)} & 0.53 (16/30) & 0.43 (13/30) & 0.33 (10/30) 
            & \multicolumn{1}{|c}{0.53} \\
            & LLM-as-Judge 
            & \multicolumn{1}{|c}{1.00 (30/30)} & 0.97 (29/30) & 1.00 (30/30) & 0.97 (29/30) & 1.00 (30/30) 
            & \multicolumn{1}{|c}{0.93 (28/30)} & 0.93 (28/30) & 0.97 (29/30) & 0.87 (26/30) 
            & \multicolumn{1}{|c}{0.96} \\
            \midrule
            \multirow{2}*{\makecell{LLaMA3.1-8B\\-Instruct}} 
            & Exact Match 
            & \multicolumn{1}{|c}{0.70 (21/30)} & 0.60 (18/30) & 0.67 (20/30) & 0.40 (12/30) & 0.33 (10/30) 
            & \multicolumn{1}{|c}{0.70 (21/30)} & 0.50 (15/30) & 0.53 (16/30) & 0.37 (11/30) 
            & \multicolumn{1}{|c}{0.53} \\
            & LLM-as-Judge 
            & \multicolumn{1}{|c}{1.00 (30/30)} & 1.00 (30/30) & 1.00 (30/30) & 0.93 (28/30) & 0.97 (29/30) 
            & \multicolumn{1}{|c}{1.00 (30/30)} & 0.90 (27/30) & 0.93 (28/30) & 0.90 (27/30) 
            & \multicolumn{1}{|c}{0.96} \\
            \midrule
            \multirow{2}*{Qwen3-8B} 
            & Exact Match 
            & \multicolumn{1}{|c}{0.83 (25/30)} & 0.73 (22/30) & 0.77 (23/30) & 0.67 (20/30) & 0.53 (16/30) 
            & \multicolumn{1}{|c}{0.80 (24/30)} & 0.70 (21/30) & 0.50 (15/30) & 0.40 (12/30) 
            & \multicolumn{1}{|c}{0.66} \\
            & LLM-as-Judge 
            & \multicolumn{1}{|c}{1.00 (30/30)} & 1.00 (30/30) & 1.00 (30/30) & 1.00 (30/30) & 0.97 (29/30) 
            & \multicolumn{1}{|c}{1.00 (30/30)} & 0.97 (29/30) & 0.93 (28/30) & 0.93 (28/30) 
            & \multicolumn{1}{|c}{0.98} \\
            \midrule
            \multirow{2}*{Qwen2.5-14B} 
            & Exact Match 
            & \multicolumn{1}{|c}{0.83 (25/30)} & 0.67 (20/30) & 0.76 (23/30) & 0.60 (18/30) & 0.57 (17/30) 
            & \multicolumn{1}{|c}{0.67 (20/30)} & 0.80 (24/30) & 0.63 (19/30) & 0.43 (13/30) 
            & \multicolumn{1}{|c}{0.66} \\
            & LLM-as-Judge 
            & \multicolumn{1}{|c}{1.00 (30/30)} & 1.00 (30/30) & 0.97 (29/30) & 1.00 (30/30) & 0.97 (29/30) 
            & \multicolumn{1}{|c}{1.00 (30/30)} & 0.97 (29/30) & 0.90 (27/30) & 0.93 (28/30) 
            & \multicolumn{1}{|c}{0.97} \\  
			\bottomrule
		\end{tabular}
	}
\end{table*}


\subsection{Further Analysis on \textit{Proprietary Language Models}}

In our main experiment, we employ DeepSeek-V3 and DeepSeek-R1 as the proprietary language models for the following roles: 

\begin{enumerate}
    \item \underline{\textit{Training Data Construction:}} As descirbed in \Cref{subsec:training_data_construction} and \Cref{fig:data_construction}, we employed the proprietary language models to first select the data that are suitable for RL from the original datasets, and then construct reference reasoning trajectories for the verification of reasoning process.
    \item \underline{\textit{Evaluation:}} During the RL process, the proprietary language model DeepSeek-V3 serves as the judge of several reward components, e.g., entity/relation coverage via knowledge graph extraction, evidence quality (EBM level) scoring, etc. It is also adopted in the \textit{LLM-as-Judge} paradigm to evaluate the correctness of models' generated responses across medical benchmarks.
\end{enumerate}

In this section, we present an in-depth analysis of the use of proprietary language models in our experimental pipeline, 
encompassing a systematic summary of their strengths, 
and human expert assessment of the judgments produced by these models.

\subsubsection{Advantages of Proprietary Language Models}

We employ the proprietary language models in our \ours pipeline for the following reasons:

\textit{\textbf{Reliability.}} 
LLM-as-Judge has become a standard evaluation approach for performance judgment, as human evaluation is costly and unscalable for open-ended tasks. This paradigm is now widely adopted in major LLM benchmarks where rule-based metrics fail to capture semantic correctness~\cite{zheng2023judging,li2025generation}. 
Specifically, this strategy outperforms traditional metrics on complex medical reasoning tasks, since medical problems exhibit key characteristics that challenge conventional evaluation:

\begin{itemize}
    \item The answers are often expressed in diverse phrasings (no fixed template).
    \item Correct responses may require justifying diagnoses or ruling out differential conditions.
    \item There are multiple semantically equivalent formulations that can all be valid.
\end{itemize}

In such settings, exact-match against gold keys severely underestimates model capability (e.g., penalizing ``myocardial infarction'' as incorrect simply because the reference says ``acute myocardial infarction''). Human evaluation, while ideal, is prohibitively expensive at scale and still subject to inter-annotator disagreement, even among clinicians. 
In contrast, LLM-as-Judge enables semantic equivalence judgment and logical coherence assessment, aligning better with the open-ended nature of clinical reasoning. 
Moreover, our evaluation process mitigates the risk of judge bias by enforcing a fixed and well-defined output format.

\textit{\textbf{Robustness.}} 
Moreover, during the training data construction phase, particularly in the data filtering process based on reasoning complexity and in generating reference reasoning trajectories for medical questions to support subsequent RL verification, we first used DeepSeek-R1 to generate multiple candidate answers, and then employed DeepSeek-V3 to evaluate the quality of these sequences generated by DeepSeek-R1, thereby enhancing fault tolerance.

\begin{table*}[th] 
    \centering
    \caption{\label{tab:failure-case}
    Failure case analysis with a held-out set of 300 questions from the MedQA-USMLE and MedMCQA datasets, where the model equipped with \ours has produced incorrect answers. 
    Here we employ \textit{Qwen2.5-14B} for experimental conduction.
    }
	\setlength{\tabcolsep}{1pt}
	\resizebox{1\linewidth}{!}{
		\begin{tabular}{c|c|c|p{9cm}}
            \toprule
            \multirow{2}{*}{Category} & \multirow{2}{*}{Sub-Category} & \multicolumn{2}{c}{Failure Case} \\
            \cline{3-4}
             & & \multicolumn{1}{c|}{Component} & \multicolumn{1}{c}{Content} \\
            \midrule
			\multirow{15}*{\makecell{Retrieval Failures\\(42\%)}} 
			& \multirow{8}*{\makecell{Irrelevant Document Retrieval\\(28\%)}} & \multirow{3}*{Question} & \textit{A 65-year-old man presents with ``sudden-onset chest pain and shortness of breath.'' Vital signs: BP 90\/60 mmHg, HR 120 bpm. Physical exam reveals bilateral pulmonary crackles. ECG shows ST-segment elevation.} \\
            \cline{3-4}
			&  & {\multirow{3}*{Model Response}} & The model retrieved a document discussing ``\textit{gastroesophageal reflux disease as a cause of non-cardiac chest pain}'' and recommended upper endoscopy. \\
            \cline{3-4}
            &  & \multirow{2}*{Error Analysis} & The term ``\textit{chest pain}'' without clear cardiac context led the retriever to off-topic documents. \\
            \cline{2-4}
            & \multirow{7}*{\makecell{High-Quality Evidence Ignored\\(14\%)}} & \multirow{2}*{Question} & \textit{Which medication is most likely to cause hypoglycemia in a patient with type 2 diabetes?} \\
            \cline{3-4}
            &  & \multirow{4}*{Model Response} & The model cited a 2020 observational study (Level IV evidence) claiming ``\textit{rosiglitazone is associated with hypoglycemia}'', while overlooking a 2023 large-scale RCT (Level I evidence) that clearly identified ``\textit{glimepiride}'' as having the highest hypoglycemia risk. \\
            \cline{3-4}
            &  & Error Analysis & The model prioritized recency over methodological rigor. \\
            \midrule
            \multirow{13}*{\makecell{Reasoning Failures\\(58\%)}} 
			& \multirow{7}*{\makecell{Misinterpretation of Evidence\\(33\%)}} & \multirow{3}*{Question} & \textit{A literature excerpt states: ``Long-term NSAID use is associated with an increased risk of gastric ulcers.'' A patient with osteoarthritis has been taking ibuprofen chronically and now presents with melena.} \\
            \cline{3-4}
            &  & \multirow{2}*{Model Response} & The model concluded that ``\textit{The melena is caused by ibuprofen-induced gastric ulcer; discontinue NSAIDs and switch to a COX-2 inhibitor.}'' \\
            \cline{3-4}
            &  & \multirow{2}*{Error Analysis} & The source only reported an association, not causation, and the model ignored other etiologies. \\
            \cline{2-4}
            & \multirow{6}*{\makecell{Incomplete Reasoning Chains\\(25\%)}} & \multirow{2}*{Question} & \textit{A 40-year-old woman presents with fever, arthralgia, and rash. Lab findings: positive ANA (1:160), low complement C3.} \\
            \cline{3-4}
            &  & \multirow{2}*{Model Response} & The model directly diagnosed ``\textit{systemic lupus erythematosus (SLE)}'' without considering alternative diagnoses. \\
            \cline{3-4}
            &  & \multirow{2}*{Error Analysis} & The model failed to evaluate drug-induced lupus or mixed connective tissue disease. \\
			\bottomrule
		\end{tabular}
	}
\end{table*}

\textit{\textbf{Reproducibility.}} 
We have also provided the prompts used during data construction and performance judgment, including those guiding content validity assessment, reasoning difficulty–based sample selection, question rewriting, reward verification in the RL process, and answer correctness evaluation, as detailed in \Cref{sec:appendix_prompts}.

\subsubsection{Ablations of the Proprietary Language Models}
\label{subsubsec:appendix_ablation_proprietary_llm}

Here we use Qwen2.5-7B as the backbone model and conduct experiments by replacing \textbf{DeepSeek-V3} (utilized in main experiments) with open-sourced alternative \textbf{LLaMA3.1-70B-Instruct}. 
We separate model dependence into \underline{\textit{a)}} trajectory scoring dependence during data construction and \underline{\textit{b)}} LLM-as-Judge dependence during evaluation. 

As shown in \Cref{tab:cross_scorer_judge}, 
we have disentangled two possible sources of dependence: the trajectory scorer used during training data construction and the final LLM-as-Judge evaluator used during evaluation. 
Even under the fully open-weight trajectory scorer and LLM-as-Judge evaluator setting, Med-R$^3$ obtains the best average score across all baselines. 
It indicates that the improvement is not tied to a specific proprietary scorer or evaluator preference, but comes from our proposed Med-R$^3$ framework.

\subsubsection{Meta-Evaluation of Proprietary Language Models}
\label{subsubsec:appendix_meta-evaluation}

Here we have conducted a small-scale validation study to evaluate the quality of proprietary language model in comparison with human-annotation (we employ DeepSeek-V3 in our main experiment) during the data curation and RL process.

\begin{itemize}
    \item \underline{\textit{Entity/Relation Extraction Accuracy} (\Cref{eq:kg_retrieval_version}):} We randomly sampled 200 reasoning steps from the training trajectories and asked three board-certified physicians to annotate entities and relations according to our schema (e.g., ``A causes B'', ``A inhibits B''). Compared to human annotations, DeepSeek-V3's extractions achieve:
    \begin{itemize}
        \item Entity: 92.3\% (F1 score)
        \item Relation (directional): 86.7\% (F1 score)
    \end{itemize}
    The inter-annotator agreement (Fleiss' k~\cite{fleiss1971measuring}) was 0.81, indicating substantial agreement.
    \item \underline{\textit{EBM Level Agreement} (\Cref{eq:authority}):} For 150 retrieved documents from the external knowledge corpus, we compared DeepSeek-V3's EBM level assignments (Level I–VI) against consensus labels from two independent clinical researchers. The weighted Cohen's k~\cite{cohen1968weighted} was 0.74, reflecting strong agreement, especially for high-quality evidence (Level I–II), where the precision exceeded 89\%.
    \item \underline{\textit{Answer Correctness Judgment}:} We have also held the human meta-evaluation study to assess the judge quality of different evaluation strategies, including \textit{exact-match} and \textit{LLM-as-Judge}. Specifically, we sample 30 instances from each medical dataset, collect the generation results from Qwen2.5-7B, LLaMA3.1-8B-Instruct, Qwen3-8B and Qwen2.5-14B, and then report the judgment correctness of \textit{exact-match} and \textit{LLM-as-Judge from the proprietary model DeepSeek-V3} (employed in our main experiment).
    As depicted in \Cref{tab:meta-analysis}, experimental results show that the proprietary model evaluator achieves approximately 97\% agreement with human experts, whereas the exact-match judge attains only about 60\% alignment on average. 
    On benchmarks related to real-world clinical scenarios, such as Rare- Arena, MedXpertQA, HealthBench, and NEJMQA, the \textit{exact-match} judge exhibits even lower alignment with human experts. It is primarily because questions in these datasets are often open-ended or admit multiple clinically valid answers, whereas exact-match evaluation only considers literal string identity and fails to account for semantic equivalence or acceptable paraphrasing in medical reasoning. This further underscores the reliability of \textit{LLM-as-Judge}.
\end{itemize}

\section{Case Studies}

\begin{figure*}[th]
    \centering
    \includegraphics[width=1\textwidth]{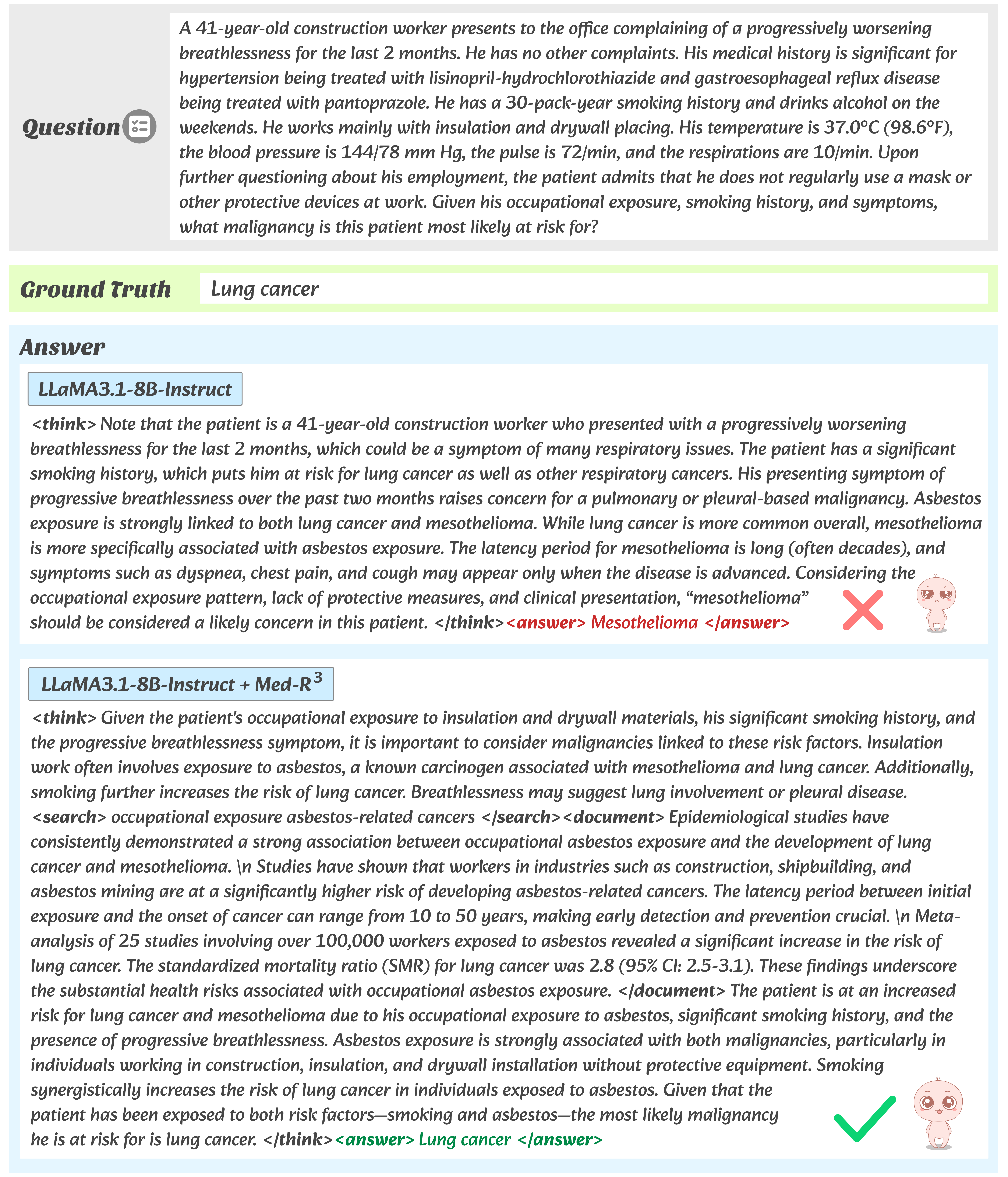}
    \caption{Case study of Med-R$^3$.}
    \label{fig:case_study}
\end{figure*}  

We have provided case studies in \Cref{fig:case_study}. Below we describe the core improvements introduced by our method:

\begin{itemize}
    \item \textit{\underline{Proactive Retrieval Triggering}}: The model doesn’t just retrieve when prompted—it identifies knowledge gaps (e.g., ``Is asbestos exposure more strongly linked to mesothelioma or lung cancer?'') and initiates search autonomously.
    \item \textit{\underline{Evidence-Based Decision Making}}: Instead of relying on memorized associations, it integrates structured evidence (e.g., SMR values, study populations) into its reasoning chain.
    \item \textit{\underline{Multi-Factor Risk Integration}}: It combines occupational history, smoking status, symptom progression, and epidemiological data to reach a balanced conclusion.
\end{itemize}

\section{Failure Case Analysis}
\label{sec:appendix_failure_analysis}

\subsection{In-Domain Analysis}
\label{subsec:ID_failure_analysis}

We have expanded our analysis of failure cases using a held-out set of 300 questions from the MedQA-USMLE and MedMCQA datasets where \ours has produced incorrect answers (the backbone model we employ here is Qwen2.5-14B, i.e., Qwen2.5-14B + Med-R$^3$). Two board-certified physicians independently categorized errors into retrieval and reasoning types, with a Fleiss' k~\cite{fleiss1971measuring} of 0.79. Our key observations are outlined in \Cref{tab:failure-case}.


The failure cases can be categorized into the following types:

\begin{itemize}
    \item \textit{\underline{Retrieval Failures}} (42\%)
    \begin{itemize}
        \item \textit{\underline{Irrelevant Document Retrieval}} (28\%): Sometimes the ambiguous symptom descriptions lead to off-topic retrieved documents.
        \item \textit{\underline{High-Quality Evidence Ignored}} (14\%): The model sometimes prioritizes recent but lower-evidence papers over systematic reviews or RCTs.
    \end{itemize}
    \item \textit{\underline{Reasoning Failures}} (58\%)
    \begin{itemize}
        \item \textit{\underline{Misinterpretation of Evidence}} (33\%): Although the model has retrieved correct documents, it confuses correlation with causation.
        \item \textit{\underline{Incomplete Reasoning Chains}} (25\%): The model gives premature conclusion without integrating multi-step clinical logic.
    \end{itemize}
\end{itemize}

\subsection{Out-of-Domain Analysis}
\label{subsec:OOD_failure_analysis}

\begin{table}[t]
    \centering
    \caption{
    \label{tab:error-analysis-OOD}
    Failure case analysis on the MedXpertQA dataset. 
    Here we employ Qwen2.5-14B + \ours for experimental conduction, and use DeepSeek-V3 for error type labeling.
    }
        \begin{tabular}{lcc}
            \toprule
            \textbf{Error Type} & \textbf{Count} & \textbf{Percentage} \\
            \midrule
            Reasoning Process Error      & 94 & 47\% \\
            Lack of Medical Knowledge    & 62 & 31\% \\
            Question Understanding Error & 36 & 18\% \\
            Formatting Error             & 8  & 4\%  \\
            \bottomrule
        \end{tabular}
\end{table}

We have also conducted error analysis on the out-of-domain (OOD) dataset MedXpertQA, the expert-level questions of which are truly challenging for all methods. 
We sampled 200 incorrectly answered questions by Qwen2.5-14B + Med-R$^3$, and use DeepSeek-V3 to label each error type, with results shown in \Cref{tab:error-analysis-OOD}.


\section{Ethical Considerations}
\label{sec:ethical_considerations}

While deploying and utilizing \ours framework holds potential in medical contexts, it also brings several ethical considerations that must be addressed to ensure responsible and beneficial use. 
The external knowledge corpora and medical data involved in retrieval ensure data anonymization, encryption, and restricted access to authorized personnel throughout the retrieval and reasoning workflows. 
Additionally, \ours adhere to healthcare regulations and ethical guidelines. Its development and deployment are overseen by medical and ethical review bodies to ensure alignment with industry standards and patient safety imperatives. 

\clearpage

\section{Prompts}
\label{sec:appendix_prompts}

Here we present the prompts used throughout our pipeline in \ours. 
Only the English version is presented due to LaTeX compilation issues with non-English languages.

\begin{tcolorbox}[breakable, title=\textbf{Prompt: Content Suitability Judgment}, colback=gray!7, colframe=gray!50!black, boxrule=1pt]
Please judge whether the following multiple-choice question is suitable for conversion into an open-ended question.
The question to be converted must meet the following conditions:

\begin{enumerate}
    \item After removing the options, the question itself remains valid, and the answer is unique and correct.
    \item There is no ambiguity in the question and answer.
    \item The question must have a unique optimal answer, not a range or a vague value.
    \item Questions with negative options, such as selecting the option that does not meet the conditions or the ``least likely'' option, are not suitable. 
    \item In other cases, please use your logical judgment to make a decision.
\end{enumerate}

\vspace{0.5cm}

\textbf{\# Question}\\
\{question\}

\vspace{0.5cm}

\textbf{\# Correct Answer}\\
\{answer\}

\vspace{0.5cm}

\textbf{\# Misleading Options}\\
\{misleading\_options\}

\vspace{0.5cm}

\hdashrule{\textwidth}{1pt}{3pt}

\vspace{0.5cm}

\textbf{Output Format:}  
\begin{verbatim}
```json
{
    "unique": True,
    "reason": "Because..."
}
'''
\end{verbatim}

\end{tcolorbox}

\begin{tcolorbox}[breakable, title=\textbf{Prompt: Reasoning Complexity Filtering}, colback=gray!7, colframe=gray!50!black, boxrule=1pt]
You are an expert in the medical field. 
You will be given a question, a student's answer, the correct and unique answer to the question, and other misleading options.
Please compare the student's answer with the correct answer and analyze whether the student's answer is correct.
In the student's answer, the \texttt{<think>}...\texttt{</think>} tag wraps the thinking process, and the \texttt{<answer>}...\texttt{</answer>} tag wraps the final answer.
During the model's reasoning process, uncertain parts are encapsulated using the \texttt{<search>}\textit{search query}\texttt{</search>} tags to facilitate subsequent information retrieval. Once relevant content is retrieved, the model continues its reasoning based on the retrieved information.
It is important to note that the content within \texttt{<document>} \textit{search results} \texttt{</document>} represents externally retrieved information and is not generated by the model itself. Therefore, this content should not be used to assess the model's logical reasoning ability. Instead, it should be used solely to evaluate the model's capability to process and integrate external information. \\

Please judge the student's thinking process and answer based on the correct answer, rate it from 1 to 5 points, and explain the reason. \\

\textbf{\textit{\underline{5 Point Answer Criteria}}}: 
\begin{enumerate}
    \item The thinking process is logical and seamless, and the reasoning process is specific and clear.
    \item The final answer is consistent with the correct answer, allowing synonyms, abbreviations, etc. of the correct answer, but cannot contain incorrect options.
\end{enumerate}

\textbf{\textit{\underline{3-4 Point Answer Criteria}}}: 
\begin{enumerate}
    \item The thinking process is reasonable. 
    \item The final answer is consistent with the correct answer, allowing some supplements, as long as they do not conflict with the correct answer, and the correct answer is the main one, not other misleading options.
\end{enumerate}

\textbf{\textit{\underline{1-2 Point Answer Criteria}}}: 
\begin{enumerate}
    \item The thinking process is not clear.
    \item The final answer is inconsistent with the correct answer or contains incorrect options.
    \item Contains garbled characters, format errors, disorder, and irrelevant information.
\end{enumerate}

\vspace{0.5cm}

\textbf{\# Question}\\
\{question\}

\vspace{0.5cm}

\textbf{\# Student's Answer}\\
\{answer\}

\vspace{0.5cm}

\textbf{\# Correct Answer}\\
\{correct\_answer\}

\vspace{0.5cm}

\textbf{\# Misleading Options}\\
\{misleading\_options\}

\hdashrule{\textwidth}{1pt}{3pt}

\vspace{0.5cm}

\textbf{Output Format:}  
\begin{verbatim}
```json
{
    "score": xxx,
    "reason": "..."
}
'''
\end{verbatim}

\end{tcolorbox}

\begin{tcolorbox}[breakable, title=\textbf{Prompt: Open-Ended Standardization}, colback=gray!7, colframe=gray!50!black, boxrule=1pt]
You are an expert in question reformulation within the medical field. 
Please convert the following multiple-choice question into an open-ended question. 
Please try to keep the content of the question unchanged as much as possible, and modify the last question into an open-ended inquiry, that is, modify the original ``Which of the following is'', ``The most likely option is'', etc. 
The modified question should also be semantically smooth and unambiguous, and the answer to the question should be consistent with the correct answer to the original question.
The language used in the modified question should be consistent with the original question.

\vspace{0.5cm}

\textbf{\# Original Question}\\
\{question\}

\vspace{0.5cm}

\textbf{\# Answer}\\
\{answer\}

\vspace{0.5cm}

\hdashrule{\textwidth}{1pt}{3pt}

\vspace{0.5cm}

\textbf{Output Format:}  
\begin{verbatim}
```json
{
    "question": "..."
}
'''
\end{verbatim}

\end{tcolorbox}

\begin{tcolorbox}[breakable, title=\textbf{Prompt: Reasoning Process Generation (with Retrieval, for training data construction, rollout in Med-R$^3$, and inference mode of \textit{CoT-RAG})}, colback=gray!7, colframe=gray!50!black, boxrule=1pt]
You are a medical expert. 
Given a question, you should answer it by first thinking about the reasoning process in the mind and then providing the final answer. 
Please answer the question in the format of \texttt{<think>}...\texttt{</think><answer>}...\texttt{</answer>}. 
That is, \texttt{<think>}\textit{Here is the reasoning process}\texttt{</think>}\texttt{<answer>}\textit{answer}\texttt{</answer>}. 
You should perform thinking with decomposing, reflecting, brainstorming, verifying, refining, and revising. 
Besides, you can perform searching for uncertain knowledge if necessary with the format of \texttt{<search>}\textit{search query}\texttt{</search>} during your thinking process.
Then, the search system will provide you with the retrieval information with the format of \texttt{<document>} \textit{search results} \texttt{</document>}.
The answer needs to summarize the reasoning process and give the final answer.
You are required to continue your reasoning and response in conjunction with the existing answer and retrieved information, ensuring that the subsequent answers you generate maintain coherence with the previously generated answers.

\vspace{0.5cm}

\textbf{\# Question}\\
\{reformulated\_question\}

\vspace{0.5cm}

\textbf{\# Existing Answer and Retrieved Information}\\
\{existing\_answer\} [optional]

\end{tcolorbox}

\begin{tcolorbox}[breakable, title=\textbf{Prompt: Reasoning Process Generation \\ (without Retrieval, for inference mode of \textit{CoT})}, colback=gray!7, colframe=gray!50!black, boxrule=1pt]
You are a medical expert. 
Given a question, you should answer it by first thinking about the reasoning process in the mind and then providing the final answer. 
Please answer the question in the format of \texttt{<think>}...\texttt{</think><answer>}...\texttt{</answer>}. 
That is, \texttt{<think>}\textit{Here is the reasoning process}\texttt{</think>}\texttt{<answer>}\textit{answer}\texttt{</answer>}. 
You should perform thinking with decomposing, reflecting, brainstorming, verifying, refining, and revising. 
The answer needs to summarize the reasoning process and give the final answer.

\vspace{0.5cm}

\textbf{\# Question}\\
\{reformulated\_question\}

\end{tcolorbox}

\begin{tcolorbox}[breakable, title=\textbf{Prompt: Answer Generation \\ (without Retrieval, for inference mode of \textit{Naive})}, colback=gray!7, colframe=gray!50!black, boxrule=1pt]
You are a medical expert, please provide answer for the following question. 

\vspace{0.5cm}

\textbf{\# Question}\\
\{reformulated\_question\}

\end{tcolorbox}

\begin{tcolorbox}[breakable, title=\textbf{Prompt: Answer Generation \\ (with Retrieval, for inference mode of \textit{RAG})}, colback=gray!7, colframe=gray!50!black, boxrule=1pt]
You are a medical expert. 
Given the following question, please consult the retrieved documents, identify key information that are directly related to the question, and provide the answer. 

\vspace{0.5cm}

\textbf{\# Question}\\
\{reformulated\_question\}

\vspace{0.5cm}

\textbf{\# Retrieved Documents}\\
\{retrieved\_documents\}

\end{tcolorbox}

\begin{tcolorbox}[breakable, title=\textbf{Prompt: Medical Knowledge Graph Extraction}, colback=gray!7, colframe=gray!50!black, boxrule=1pt]
You are a medical expert. 
Given a reasoning process for solving a medical problem in the format of \texttt{<think>}...\texttt{</think>} and \texttt{<answer>}...\texttt{</answer>}. The content within \texttt{<think>}...\texttt{</think>} demonstrates the thought process and may use \texttt{<search>}\textit{search query}\texttt{</search>} to mark uncertain knowledge that requires searching. The search system provides relevant information in the format of \texttt{<document>} \textit{search results} \texttt{</document>}. 
Your task is to extract important medical concepts, relationships, and attributes from the given reasoning process and represent them in the knowledge graph format.

\vspace{0.5cm}

\textbf{\# Reasoning process}\\
\{reasoning\_reference\}

\vspace{0.5cm}

\hdashrule{\textwidth}{1pt}{3pt}

\vspace{0.5cm}

\textbf{Output Format:}  
\begin{verbatim}
```json
[[
    "entity1", "relationship", 
    "entity2", "if_retrieval"]
], ...]
'''
\end{verbatim}

\vspace{0.5cm}

Definitions:
\begin{enumerate}
    \item \textbf{entity1, relationship, entity2}: Clearly extract entities or relationships from the natural language (e.g., ``patient'', ``has symptom'', ``fever''). 
    \item \textbf{if\_retrieval}: A boolean value (\texttt{1/0}) indicating whether the entity or relationship is retrieved from the search results within \texttt{<document>}...\texttt{</document>}.
\end{enumerate}

\vspace{0.5cm}

Processing rules:
\begin{enumerate}
    \item \textbf{\textit{\underline{Overall}}}: Extract only entities and relationships related to solving medical problems, ignoring irrelevant background information. The knowledge graph should be concise and clear, avoiding redundancy, and based on facts, avoiding subjective speculation. 
    \item \textbf{\textit{\underline{Entity Extraction}}}: Must be specific medical concepts (diseases, symptoms, drugs, etc.), excluding vague descriptions (such as ``some conditions'' or ``related factors''). Synonymous expressions should be merged (e.g., ``myocardial infarction'' instead of ``heart attack''). 
    \item \textbf{\textit{\underline{Relationship Definition}}}: Use verb phrases (cause, inhibit, lead to, accompany, etc.) and must have a clear directionality (A→B or B→A should conform to medical logic). Negative relationships should be explicitly marked (e.g., ``does not cause'', ``rule out''). 
    \item \textbf{\textit{\underline{Retrieval Marking Determination}}}: If an entity or relationship is directly from the \texttt{<document>} \textit{search results} \texttt{</document>}, mark ``\texttt{if\_retrieval=1}''; otherwise, ``\texttt{if\_retrieval=0}''. The search keywords themselves should not be considered as entities or relationships. 
    \item \textbf{\textit{\underline{Special Handling}}}: Retain hypothetical relationships in the reasoning process (marked as [hypothesis]), convert time relationships to medical temporal expressions (acute/chronic/ongoing period, etc.), and quantify probabilistic conclusions as (high/medium/low) risk levels.
\end{enumerate}

\end{tcolorbox}

\begin{tcolorbox}[breakable, title=\textbf{Prompt: Evidence Quality Judgment}, colback=gray!7, colframe=gray!50!black, boxrule=1pt]
You are an expert in evidence quality annotation within the medical field. There are 6 quality levels of evidence, ranging from the highest to the lowest as follows: Meta-Analyses and Systematic Reviews, Randomized Controlled Trials, Cohort Studies, Case-Control Studies, Individual Case Series or Case Reports, Background Information or Expert Opinion. Please classify the following evidence document based on its structure and characteristics, providing only the names of the levels, without any additional description:

\vspace{0.5cm}

\textbf{\# Evidence}\\
\{retrieved\_document\}

\end{tcolorbox}

\begin{tcolorbox}[breakable, title=\textbf{Prompt: Answer Correctness Judgment}, colback=gray!7, colframe=gray!50!black, boxrule=1pt]
You are an expert in the medical field. 
You will be given a question, a student's answer, and the correct answer to the question.
Please compare the student's answer with the correct answer and analyze whether the student's answer is correct.
In the student's answer, the \texttt{<think>}...\texttt{</think>} tag wraps the thinking process, and the \texttt{<answer>}...\texttt{</answer>} tag wraps the final answer. 
Please judge the student's answer encapsulated within the \texttt{<answer>}...\texttt{</answer>} tag based on the correct answer, rate it from 0 to 2 points, and explain the reason. \\

\textbf{\textit{\underline{2 Point Answer Criteria}}}: \\
The final answer is consistent with the correct answer, allowing synonyms, abbreviations, etc. \\

\textbf{\textit{\underline{1 Point Answer Criteria}}}: \\
The final answer is consistent with the correct answer, allowing some supplements, as long as they do not conflict with the correct answer. \\

\textbf{\textit{\underline{0 Point Answer Criteria}}}: \\
The final answer is inconsistent with the correct answer, or contains garbled characters, format errors, disorder, and irrelevant information.

\vspace{0.5cm}

\textbf{\# Question}\\
\{question\}

\vspace{0.5cm}

\textbf{\# Student's Answer}\\
\{answer\}

\vspace{0.5cm}

\textbf{\# Correct Answer}\\
\{correct\_answer\}

\hdashrule{\textwidth}{1pt}{3pt}

\vspace{0.5cm}

\textbf{Output Format:}  
\begin{verbatim}
```json
{
    "score": xxx,
    "reason": "..."
}
'''
\end{verbatim}

\end{tcolorbox}





\end{document}